\documentclass[journal,10pt]{IEEEtran}

\usepackage{algorithm,algorithmic}
\usepackage{comment}

   \usepackage[pdftex]{graphicx}
  \graphicspath{{Figures/}}
\usepackage{multicol, blindtext}
\usepackage{amsmath,amsthm}
\usepackage{enumerate}
\usepackage{amsfonts}
\title{Robust SAR STAP via Kronecker Decomposition                    }

\newtheorem{theorem}{Theorem}[section]

\begin{document}
\author{Kristjan~Greenewald,~\IEEEmembership{Student Member,~IEEE,} Edmund~Zelnio, and~Alfred~Hero III,~\IEEEmembership{Fellow,~IEEE}

\thanks{K. Greenewald and A. Hero III are with the Department
of Electrical Engineering and Computer Science, University of Michigan, Ann Arbor,
MI, USA. E. Zelnio is with the Air Force Research Laboratory, Wright Patterson Air Force Base, OH 45433, USA. This research was partially supported by the Dept. of Air Force grant FA8650-15-D-1845, AFOSR grant FA8650-07-D-1220-0006, and ARO MURI grant W911NF-11-1-0391. Approved for public release, PA Approval \#88ABW-2014-6099.
}
}
%

\iffalse
\includecomment{LongerTheorems}
\excludecomment{ShorterTheorems}
\includecomment{ThmApp}
\includecomment{CD}
\includecomment{CD1}
\includecomment{CD2}
\includecomment{CD3}

\else
\excludecomment{LongerTheorems}
\includecomment{ShorterTheorems}
\excludecomment{ThmApp}
\excludecomment{CD}
\excludecomment{CD1}
\excludecomment{CD2}
\excludecomment{CD3}

\fi

 \maketitle
\begin{abstract}

This paper proposes a spatio-temporal decomposition for the detection of moving targets in multiantenna SAR. As a high resolution radar imaging modality, SAR detects and localizes non-moving targets accurately, giving it an advantage over lower resolution GMTI radars. Moving target detection is more challenging due to target smearing and masking by clutter.  Space-time adaptive processing (STAP) is often used to remove the stationary clutter and enhance the moving targets. In this work, it is shown that the performance of STAP can be improved by modeling the clutter covariance as a space vs. time Kronecker product with low rank factors. Based on this model, a low-rank Kronecker product covariance estimation algorithm is proposed, and a novel separable clutter cancelation filter based on the Kronecker covariance estimate is introduced. The proposed method provides orders of magnitude reduction in the required number of training samples, as well as improved robustness to corruption of the training data. Simulation results and experiments using the Gotcha SAR GMTI challenge dataset are presented that confirm the advantages of our approach relative to existing techniques.

\end{abstract}


\section{Introduction}


\IEEEPARstart{T}{h}e detection (and tracking) of moving objects is an important task for scene understanding, as motion often indicates human related activity \cite{newstadt2013moving}. Radar sensors are uniquely suited for this task, as object motion can be discriminated via the Doppler effect. In this work, we propose a spatio-temporal decomposition method of detecting ground based moving objects in airborne Synthetic Aperture Radar (SAR) imagery, also known as SAR GMTI (SAR Ground Moving Target Indication).

Radar moving target detection modalities include MTI radars \cite{newstadt2013moving,ender1999space}, which use a low carrier frequency and high pulse repetition frequency to directly detect Doppler shifts. This approach has significant disadvantages, however, including low spatial resolution, small imaging field of view, and the inability to detect stationary or slowly moving targets. The latter deficiency means that objects that move, stop, and then move are often lost by a tracker.

SAR, on the other hand, typically has extremely high spatial resolution and can be used to image very large areas, e.g. multiple square miles in the Gotcha data collection \cite{GotchaData}. As a result, stationary and slowly moving objects are easily detected and located \cite{ender1999space,newstadt2013moving}. Doppler, however, causes smearing and azimuth displacement of moving objects \cite{jao2001theory}, making them difficult to detect when surrounded by stationary clutter. Increasing the number of pulses (integration time) simply increases the amount of smearing instead of improving detectability \cite{jao2001theory}. Several methods have thus been developed for detecting and potentially refocusing \cite{cristallini2013efficient,cerutti2012optimum} moving targets in clutter. Our goal is to remove the disadvantages of MTI and SAR by combining their strengths (the ability to detect Doppler shifts and high spatial resolution) using space time adaptive processing (STAP) with a novel Kronecker product spatio-temporal covariance model, as explained below.  


SAR systems can either be single channel (standard single antenna system) or multichannel. Standard approaches for the single channel scenario include autofocusing \cite{fienup2001detecting} and velocity filters. Autofocusing works only in low clutter, however, since it may focus the clutter instead of the moving target \cite{fienup2001detecting,newstadt2013moving}. Velocity filterbank approaches used in track-before-detect processing \cite{jao2001theory} involve searching over a large velocity/acceleration space, which often makes computational complexity excessively high. Attempts to reduce the computational complexity have been proposed, e.g. via compressive sensing based dictionary approaches \cite{khwaja2011applications} and Bayesian inference \cite{newstadt2013moving}, but remain computationally intensive.

Multichannel SAR has the potential for greatly improved moving target detection performance \cite{ender1999space,newstadt2013moving}. Standard multiple channel configurations include spatially separated arrays of antennas, flying multiple passes (change detection), using multiple polarizations, or combinations thereof \cite{newstadt2013moving}. 


\subsection{Previous Multichannel Approaches}
Several techniques exist for using multiple radar channels (antennas) to separate the moving targets from the stationary background. SAR GMTI systems have an antenna configuration such that each antenna transmits and receives from approximately the same location but at slightly different times \cite{GotchaData,ender1999space,newstadt2013moving,cerutti2012optimum}. Along track interferometry (ATI) and displaced phase center array (DPCA) are two classical approaches \cite{newstadt2013moving} for detecting moving targets in SAR GMTI data, both of which are applicable only to the two channel scenario. Both ATI and DPCA first form two SAR images, each image formed using the signal from one of the antennas. To detect the moving targets, ATI thresholds the phase difference between the images and DPCA thresholds the magnitude of the difference. A Bayesian approach using a parametric cross channel covariance generalizing ATI/DPCA to $p$ channels was developed in \cite{newstadt2013moving}, and a unstructured method fusing STAP and a test statistic in \cite{cerutti2012optimum}. Space-time Adaptive Processing (STAP) learns a spatio-temporal covariance from clutter training data, and uses these correlations to filter out the stationary clutter while preserving the moving target returns \cite{ender1999space,ginolhac2014exploiting,klemm2002principles}. 

A second configuration, typically used in classical GMTI, uses phase coherent processing of the signals output by an antenna array for which each antenna receives spatial reflections of the same transmission at the same time. This contrasts with the above configuration where each antenna receives signals from different transmissions at different times. In this second approach the array is designed such that returns from different angles create different phase differences across the antennas \cite{klemm2002principles,ginolhac2014exploiting,rangaswamy2004robust,kirsteins1994adaptive,haimovich1996eigencanceler,conte2003exploiting}. In this case, the covariance-based STAP approach, described above, can be applied to cancel the clutter \cite{rangaswamy2004robust,ginolhac2014exploiting,haimovich1996eigencanceler}. 

In this paper, we focus on the first (SAR GMTI) configuration and propose a covariance-based STAP algorithm with a customized Kronecker product covariance structure. The SAR GMTI receiver consists of an array of $p$ phase centers (antennas) processing $q$ pulses in a coherent processing interval. Define the array $\mathbf{X}^{(m)} \in \mathbb{C}^{p\times q}$ such that $X_{ij}^{(m)}$ is the radar return from the $j$th pulse of the $i$th channel in the $m$th range bin. Let $\mathbf{x}_m = \mathrm{vec}(\mathbf{X}^{(m)})$. The target-free radar data $\mathbf{x}_m$ is complex valued and is assumed to have zero mean. Define
\begin{align}
\mathbf{\Sigma} = \mathrm{Cov}[\mathbf{x}] = E[\mathbf{x} \mathbf{x}^H].
\end{align}

The training samples, denoted as the set $\mathcal{S}$, used to estimate the SAR covariance $\mathbf{\Sigma}$ are collected from $n$ representative range bins. 
The standard sample covariance matrix (SCM) is given by
\begin{align}
\label{Eq:SCM}
\mathbf{S} = \frac{1}{n}\sum_{m\in\mathcal{S}} \mathbf{x}_m \mathbf{x}_m^H.
\end{align}
If $n$ is small, $\mathbf{S}$ may be rank deficient or ill-conditioned \cite{newstadt2013moving,ginolhac2014exploiting,greenewaldArxiv,greenewaldSSP2014}, and it can be shown that using the SCM directly for STAP requires a number $n$ of training samples that is at least twice the dimension $pq$ of $\mathbf{S}$ \cite{reed1974rapid}. In this data rich case, STAP performs well \cite{newstadt2013moving,ender1999space,ginolhac2014exploiting}. However, with $p$ antennas and $q$ time samples (pulses), the dimension $pq$ of the covariance is often very large, making it difficult to obtain a sufficient number of target-free training samples. This so-called ``small $n$ large $pq$" problem leads to severe instability and overfitting errors, compromising STAP tracking performance.


By introducing structure and/or sparsity into the covariance matrix, the number of parameters and the number of samples required to estimate them can be reduced. 
As the spatiotemporal clutter covariance $\mathbf{\Sigma}$ is low rank \cite{brennan1992subclutter,ginolhac2014exploiting,rangaswamy2004robust,ender1999space}, Low Rank STAP (LR-STAP) clutter cancelation estimates a low rank clutter subspace from $\mathbf{S}$ and uses it to estimate and remove the rank $r$ clutter component in the data \cite{bazi2005unsupervised,ginolhac2014exploiting}, reducing the number of parameters from $O(p^2q^2)$ to $O(rpq)$. 
Efficient algorithms, including some involving subspace tracking, have been proposed \cite{belkacemi2006fast,shen2009reduced}. Other methods adding structural constraints such as persymmetry \cite{ginolhac2014exploiting,conte2003exploiting}, and robustification to outliers either via exploitation of the SIRV model \cite{ginolhac2009spatio} or adaptive weighting of the training data \cite{gerlach2011robust} have been proposed. Fast approaches based on techniques such as Krylov subspace methods \cite{goldstein1998multistage,honig2002adaptive,pados2007short,scharf2008subspace} and adaptive filtering \cite{rui2011reduced,rui2010reduced} exist. All of these techniques remain sensitive to outlier or moving target corruption of the training data, and generally still require large training sample sizes \cite{newstadt2013moving}. 


Instead, for SAR GMTI we propose to exploit the explicit space-time arrangement of the covariance by modeling the clutter covariance matrix $\mathbf{\Sigma}_c$ as the Kronecker product of two smaller matrices 
\begin{equation}
\label{KronApprox}
\mathbf{\Sigma}_c = \mathbf{A}\otimes \mathbf{B},
\end{equation}
where $\mathbf{A} \in \mathbb{C}^{p\times p}$ is rank 1 and $\mathbf{B}\in \mathbb{C}^{q\times q}$ is low rank.
In this setting, the $\mathbf{B}$ matrix is the ``temporal (pulse) covariance" and $\mathbf{A}$ is the ``spatial (antenna) covariance," both determined up to a multiplicative constant. We note that this model is not appropriate for classical GMTI STAP, since that configuration the covariance has a different spatio-temporal structure that is not separable. 

Both ATI and DPCA in effect attempt to filter deterministic estimates of $\mathbf{A}$ to remove the clutter, and the Bayesian method \cite{newstadt2013moving} uses a form of this model and incorporates the matrix $\mathbf{A}$ in a hierarchical clutter model. Standard SAR GMTI STAP approaches and the method of \cite{cerutti2012optimum} do not exploit this structure when estimating the spatiotemporal covariance. To our knowledge, this work is the first to exploit spatio-temporal structure to estimate a full low-rank spatio-temporal clutter covariance.

Kronecker product covariances arise in a variety of applications (e.g. see \cite{werner2007estimation,  tsiligkaridis2013convergence,allen2010transposable,bonilla2008multi,yin2012model}). A rich set of algorithms and associated guarantees on the reduction in the number of training samples exist for estimation of covariances in Kronecker product form, including iterative maximum likelihood \cite{werner2008estimation,tsiligkaridis2013convergence}, noniterative L2 based approaches \cite{werner2008estimation}, sparsity promoting methods \cite{tsiligkaridis2013convergence,zhou2014gemini}, and robust ML SIRV based methods \cite{greenewaldSSP2014}. 


We propose an iterative L2 based algorithm to estimate the low rank Kronecker factors from the observed sample covariance. 
Theoretical results indicate significantly fewer training samples are required, and that the proposed approach improves robustness to corrupted training data. Critically, robustness allows significant numbers of moving targets to remain in the training set. We then introduce the Kron STAP filter, which projects away both the spatial and temporal clutter subspaces, thereby achieving improved cancelation. 

To summarize, the main contributions of this paper are: 1) the exploitation of the inherent Kronecker product spatio-temporal structure of the clutter covariance; 2) the introduction of the low rank Kronecker product based Kron STAP filter; 3) an algorithm for estimating the spatial and temporal clutter subspaces that is highly robust to outliers due to the additional Kronecker product structure; and 4) theoretical results demonstrating improved signal-to-interference-plus-noise-ratio\begin{CD1}
; and 5) an extension to multipass STAP
\end{CD1}
.

The remainder of the paper is organized as follows. Section \ref{Sec:Model} presents the multichannel SIRV radar model. Our low rank Kronecker product covariance estimation algorithm and our proposed STAP filter are presented in Section \ref{Sec:KSTAP}\begin{CD2}
with an extension to the case of moving target detection with multiple passes
\end{CD2}
.
Section \ref{Sec:Pred} gives theoretical performance guarantees and Section \ref{Sec:Results} gives simulation results and applies our algorithms to the Gotcha dataset. 

We denote vectors as lower case bold letters, matrices as upper case bold letters, the complex conjugate as $a^*$, the matrix Hermitian as $\mathbf{A}^H$, and the Hadamard (elementwise) product as $\mathbf{A}\odot \mathbf{B}$.

\section{SIRV Data Model}
\label{Sec:Model}
Let $\mathbf{X}\in \mathbb{C}^{p \times q}$ be an array of radar returns from an observed range bin across $p$ channels and $q$ pulses. We model $\mathbf{x} = \mathrm{vec}(\mathbf{X})$ as a spherically invariant random vector (SIRV) with the following decomposition \cite{yao1973representation,rangaswamy2004robust,ginolhac2014exploiting,ginholhac2013performance}: 
\begin{align}
\label{Eq:decomp}
\mathbf{x} = \mathbf{x}_{target} + \mathbf{x}_{clutter} + \mathbf{x}_{noise} = \mathbf{x}_{target} + \mathbf{n},
\end{align}
where $\mathbf{x}_{noise}$ is Gaussian sensor noise with $\mathrm{Cov}[\mathbf{x}_{noise}]=\sigma^2 \mathbf{I} \in \mathbb{C}^{pq \times pq}$ and we define $\mathbf{n} =\mathbf{x}_{clutter} + \mathbf{x}_{noise}$. The signal of interest $\mathbf{x}_{target}$ is the sum of the spatio-temporal returns from all moving objects, modeled as non-random, in the range bin. The return from the stationary clutter is given by $\mathbf{x}_{clutter}  =\tau \mathbf{c}$ where $\tau$ is a random positive scalar having arbitrary distribution, known as the \emph{texture}, and $\mathbf{c} \in \mathbb{C}^{pq}$ is a multivariate complex Gaussian distributed random vector, known as the \emph{speckle}. We define $\mathrm{Cov}[\mathbf{c}]= \mathbf{\Sigma}_c$. 
The means of the clutter and noise components of $\mathbf{x}$ are zero. The resulting clutter plus noise ($\mathbf{x}_{target} = 0$) covariance is given by
\begin{align}
\label{Eq:Cov}
\mathbf{\Sigma} = E[\mathbf{n}\mathbf{n}^H] =  E[\tau^2]\mathbf{\Sigma}_c + \sigma^2 \mathbf{I}.
\end{align}

The ideal (no calibration errors) random speckle $\mathbf{c}$ is of the form \cite{newstadt2013moving,ender1999space,cerutti2012optimum}
\begin{align}
\label{Eq:7}
\mathbf{c} = \mathbf{1}_p \otimes \tilde{\mathbf{c}},
\end{align}
where $\tilde{\mathbf{c}} \in \mathbb{C}^q$. The representation \eqref{Eq:7} follows because the antenna configuration in SAR GMTI is such that the $k$th antenna receives signals emitted at different times at approximately (but not necessarily exactly) the same point in space \cite{newstadt2013moving,GotchaData}. This is achieved by arranging the $p$ antennas in a line parallel to the flight path, and delaying the $k$th antenna's transmission until it reaches the point $x_i$ in space associated with the $i$th pulse. The representation \eqref{Eq:7} gives a clutter covariance of 
\begin{align}
\label{Eq:ClutterCov}
\mathbf{\Sigma}_c = \mathbf{11}^T \otimes \mathbf{B}, \quad \quad \mathbf{B} = E[\tilde{\mathbf{c}} \tilde{\mathbf{ c }}^H],
\end{align}
where $\mathbf{B}$ 
depends on the spatial characteristics of the clutter in the region of interest and the SAR collection geometry \cite{ender1999space}. 
While in SAR GMTI $\mathbf{B}$ is not exactly low rank, it is approximately low rank in the sense that significant energy concentration in a few principal components is observed over small regions \cite{borcea2013synthetic}.


Due to the long integration time and high cross range resolution associated with SAR, the returns from the general class of moving targets are more complicated, making simple Doppler filtering difficult. During short intervals for which targets have constant Doppler shift $f$ (proportional to the target radial velocity) within a range bin, the return has the form
\begin{align}
\label{Eq:Moving}
\mathbf{x} = \alpha \mathbf{d} = \alpha \mathbf{a}(f) \otimes \mathbf{b}(f),
\end{align}
where $\alpha$ is the target's amplitude, $\mathbf{a}(f) = [\begin{array}{cccc} 1 & e^{j2\pi \theta_1(f)} & \dots & e^{j 2 \pi \theta_{p}(f)} \end{array}]^T$, the $\theta_i$ depend on Doppler shift $f$ and the platform speed and antenna separation \cite{newstadt2013moving}, and $\mathbf{b} \in \mathbb{C}^q$ depends on the target, $f$, and its cross range path. The unit norm vector $\mathbf{d} =  \mathbf{a}(f) \otimes \mathbf{b}(f)$ is known as the \emph{steering vector}. For sufficiently large $\theta_i(f)$, $\mathbf{a}(f)^H \mathbf{1}$ will be small and the target will lie outside of the SAR clutter spatial subspace. The overall target return can be approximated as a series of constant-Doppler returns, hence the overall return should lie outside of the clutter spatial subspace.  Furthermore, as observed in \cite{fienup2001detecting}, for long integration times the return of a moving target is significantly different from that of uniform stationary clutter, implying that moving targets generally lie outside the temporal clutter subspace \cite{fienup2001detecting} as well. 

In practice, the signals from each antenna have gain and phase calibration errors that vary slowly across angle and range \cite{newstadt2013moving}, but these errors can be accurately modeled as constant over small regions \cite{newstadt2013moving}. Let the calibration error on antenna $i$ be $h_i e^{j\phi_i}$ and $\mathbf{h} = [\begin{array}{ccc} h_1 e^{j\phi_1},& \dots, & h_p e^{j\phi_p}\end{array}]$, giving an observed return $\mathbf{x}' = (\mathbf{h} \otimes \mathbf{I})\odot \mathbf{x}$ and a clutter covariance of 
\begin{align}
\label{Eq:KronCov}
\tilde{\mathbf{\Sigma}}_c  = (\mathbf{h} \mathbf{h}^H) \otimes \mathbf{B}= \mathbf{A} \otimes \mathbf{B}
\end{align}
implying that the $\mathbf{A}$ in \eqref{KronApprox} has rank one. 

\subsection{Space Time Adaptive Processing}\label{SubSec:STAP}

Let the vector $\mathbf{d}$ be a spatio-temporal ``steering vector" \cite{ginolhac2014exploiting}, that is, a matched filter for a specific target location/motion profile. For a measured array output vector $\mathbf x$  define the STAP filter output $y={\mathbf w}^T {\mathbf x}$, where  $\mathbf w$ is a vector of spatio-temporal filter coefficients. By \eqref{Eq:decomp} and \eqref{Eq:Moving} we have
\begin{align}
\label{Eq:Breakdown}
y = \mathbf{w}^H\mathbf{x} = \alpha \mathbf{w}^H \mathbf{d}+ \mathbf{w}^H \mathbf{n}.
\end{align}

The goal of STAP is to design the filter $\mathbf{w}$ such that the clutter is canceled ($\mathbf{w}^H \mathbf{n}$ is small) and the target signal is preserved ($\mathbf{w}^H \mathbf{d}$ is large).
For a given target with spatio-temporal steering vector $\mathbf{d}$, an optimal clutter cancellation filter is defined as the filter $\mathbf w$ that maximizes the SINR (signal to interference plus noise ratio), defined as the ratio of the power of the filtered signal $\alpha \mathbf{w}^H \mathbf{d}$ to the power of the filtered clutter and noise \cite{ginolhac2014exploiting}
\begin{align}
\label{Eq:SINRdef}
\mathrm{SINR}_{out} = \frac{|\alpha|^2 |\mathbf{w}^H\mathbf{d}|^2}{E[\mathbf{w}^H \mathbf{n}\mathbf{n}^H \mathbf{w}]} = \frac{|\alpha|^2 |\mathbf{w}^H\mathbf{d}|^2}{\mathbf{w}^H \mathbf{\Sigma} \mathbf{w}},
\end{align}
where $\mathbf{\Sigma}$ is the clutter plus noise covariance in \eqref{Eq:Cov}. 

It can be shown \cite{ender1999space,ginolhac2014exploiting} that, if the clutter covariance is known, under the SIRV model the optimal  filter for targets at locations and velocities corresponding to the steering vector $\mathbf{d}$ is given by the filter
\begin{align}
\label{Eq:opt}
\mathbf{w} = \mathbf{F}_{opt}\mathbf{d},\quad \mathbf{F}_{opt} = \mathbf{\Sigma}^{-1}.
\end{align}
Since the true covariance is unknown, we consider filters of the form
\begin{align}
\label{Eq:GenFilt}
\mathbf{w} = \mathbf{F}\mathbf{d},
\end{align}
and use the measurements to learn an estimate of the best $\mathbf F$. 




For both classical GMTI radars and SAR GMTI, the clutter covariance has low rank $r$ \cite{brennan1992subclutter,newstadt2013moving,ender1999space}. Clutter subspace processing finds a \emph{clutter subspace} $\{\mathbf{u}_i\}_{i=1}^r$ using the span of the top $r$ principal components of the clutter sample covariance \cite{ender1999space,ginolhac2014exploiting}. This gives a clutter cancelation filter $\mathbf{F}$ that projects onto the space orthogonal to the estimated clutter subspace:
\begin{align}
\label{Eq:RegStap}
\mathbf{F}= \mathbf{I} - \sum_{i=1}^r \mathbf{u}_i \mathbf{u}_i^H.
\end{align}

Since the sample covariance requires a relatively large number of training samples, obtaining sufficient numbers of target free training samples is difficult in practice \cite{newstadt2013moving,ginolhac2014exploiting}. In addition, if low amplitude moving targets are accidentally included in training, the sample covariance will be corrupted. In this case the resulting filter will partially cancel moving targets as well as clutter, which is especially problematic in online STAP implementations \cite{newstadt2013moving,belkacemi2006fast}. The proposed Kronecker STAP approach discussed below mitigates these problems as it directly takes advantage of the inherent space vs. time product structure of the clutter covariance $\mathbf{\Sigma}_c$.

\section{Kronecker STAP}
\label{Sec:KSTAP}
\subsection{Kronecker Subspace Estimation}
\label{Sec:Alg}

In this section we develop a subspace estimation algorithm that accounts for spatio-temporal covariance structure and has low computational complexity. In a high-dimensional setting, performing maximum likelihood on low-rank Kronecker product covariance estimation is computationally intensive under the Gaussian model or its SIRV extensions, and existing approximations combining Kronecker products with Tyler's estimator \cite{greenewaldSSP2014} do not give low rank estimates.

Similarly to the constrained least squares approaches of \cite{werner2008estimation,greenewaldArxiv,tsiliArxiv,greenewaldSSP2014}, we fit a low rank Kronecker product model to the sample covariance matrix ${\mathbf{ S}}$. Specifically, we minimize the Frobenius norm of the residual errors in the approximation of $\mathbf S$ by the low rank Kronecker model \eqref{Eq:KronCov}, subject to $\mathrm{rank}(\mathbf{A}) \leq r_a, \mathrm{rank}(\mathbf{B}) \leq r_b$, where the goal is to estimate $E[\tau^2] \mathbf{\Sigma}_c$. The optimal estimates of the Kronecker matrix factors $\mathbf{A}$ and $ \mathbf{B}$ in \eqref{Eq:KronCov} are given by
\begin{equation}
\label{Eq:SparseOpt}
\hat{\mathbf{A}},\hat{\mathbf{B}} = \arg\min_{\mathrm{rank}({\mathbf{A}}) \leq r_a,\mathrm{rank}({\mathbf{B}}) \leq r_b}\| \mathbf{S}-{ \mathbf{A}}\otimes{\mathbf{B}}\|_F^2.
\end{equation}

The minimization \eqref{Eq:SparseOpt} will be simplified by using the patterned block structure of $\mathbf A \otimes \mathbf B$. In particular, for a $pq \times pq$ matrix $\mathbf{M}$, define $\{\mathbf{M}(i,j)\}_{i,j=1}^p$ to be its $q \times q$ block submatrices, i.e. $\mathbf{M}(i,j) = [\mathbf{M}]_{(i-1)q + 1:iq,(j-1)q+1:jq}$. Also, let $\overline{\mathbf{M}} = \mathbf{K}_{p,q}^T \mathbf{M} \mathbf{K}_{p,q}$ where $\mathbf{K}_{p,q}$ is the $pq \times pq$ permutation operator such that $\mathbf{K}_{p,q} \mathrm{vec}(\mathbf{N}) = \mathrm{vec}(\mathbf{N}^T)$ for any $p\times q$ matrix $\mathbf{N}$.

The invertible Pitsianis-VanLoan rearrangement operator $\mathcal{R}(\cdot)$ maps $pq\times pq$ matrices to $p^2 \times q^2$ matrices and, as defined in \cite{tsiliArxiv,werner2008estimation} sets the $(i-1)p + j$th row of $\mathcal{R}(\mathbf{M})$ equal to $\mathrm{vec}(\mathbf{M}(i,j))^T$, i.e. 
\begin{align}
\label{Eq:SVD}
\mathcal{R}(\mathbf{M}) &= [\begin{array}{ccc} \mathbf{m}_1 & \dots & \mathbf{m}_{p^2}\end{array}]^T,\\\nonumber
\mathbf{m}_{(i-1)p+j} &= \mathrm{vec}(\mathbf{M}(i,j)), \quad i,j = 1,\dots,p.
\end{align}

The unconstrained (i.e. $r_a = p, r_b = q$) objective in \eqref{Eq:SparseOpt} is shown in \cite{werner2008estimation,tsiliArxiv,greenewaldArxiv} to be equivalent to a rearranged rank-one approximation problem, with a global minimizer given by
\begin{equation}
\label{Eq:Werner}
\hat{\mathbf{A}}\otimes \hat{\mathbf{B}} = \mathcal{R}^{-1}(\sigma_1 \mathbf{u}_1 \mathbf{v}_1^H),
\end{equation}
where $\sigma_1 \mathbf{u}_1 \mathbf{v}_1^H$ is the first singular component of $\mathcal{R}(\mathbf{S})$. The operator $\mathcal{R}^{-1}$ is the inverse of $\mathcal{R}$, given by
\begin{align}
\mathcal{R}^{-1}(\mathbf{M}) &= \mathbf{N} \in \mathbb{C}^{pq \times pq},\\\nonumber
\mathbf{N}(i,j) &= \mathrm{vec}^{-1}_{q,q}((\mathbf{M}_{(i-1)p + j,1:q^2})^T), \quad i,j = 1,\dots, p,
\end{align}
where $\mathrm{vec}^{-1}_{q,q}(\cdot)$ is the inverse of the vectorization operator on $q\times q $ matrices, i.e. if $\mathbf{m} = \mathrm{vec}(\mathbf{M})\in \mathbb{C}^{q\times q}$, $\mathbf{M} = \mathrm{vec}^{-1}_{q,q}(\mathbf{m})$.

When the low rank constraints are introduced, there is no closed-form solution of \eqref{Eq:SparseOpt}. An iterative alternating minimization algorithm is derived in Appendix \ref{App:LRKron} and is summarized by Algorithm \ref{alg:LRKron}. In Algorithm \ref{alg:LRKron}, $\mathrm{EIG}_{r}(\mathbf{M})$ denotes the matrix obtained by truncating the Hermitian matrix $\mathbf M$ to its first $r$ principal components, i.e.
\begin{equation}
\mathrm{EIG}_r(\mathbf{M}) := \sum_{i=1}^r \sigma_i \mathbf{u}_i \mathbf{u}_i^H,
\end{equation}
where $\sum_i \sigma_i \mathbf{u}_i \mathbf{u}_i^H$ is the eigendecomposition of $\mathbf{M}$, and the (real and positive) eigenvalues $\sigma_i$ are indexed in order of decreasing magnitude. 

The objective \eqref{Eq:SparseOpt} is not convex, but since it is an alternating minimization algorithm, it can be shown (Appendix \ref{App:LRKron}) that Algorithm \ref{alg:LRKron} will monotonically decrease the objective at each step, and that convergence of the estimates $\mathbf{A}_k,\mathbf{B}_k$ to a stationary point of the objective is guaranteed. We initialize LR-Kron with either $\hat{\mathbf{A}},\hat{\mathbf{B}}$ from the unconstrained estimate \eqref{Eq:Werner}. Monotonic convergence then guarantees that LR-Kron improves on this simple closed form estimator. 


We call Algorithm \ref{alg:LRKron} low rank Kronecker product covariance estimation, or LR-Kron. In Appendix \ref{App:LRKron} it is shown that when the initialization is positive semidefinite Hermitian the LR-Kron estimator $\hat{\mathbf{A}}\otimes \hat{\mathbf{B}}$ is positive semidefinite Hermitian and is thus a valid covariance matrix of rank $r_a r_b$.

\begin{algorithm}[H]
\caption{LR-Kron Covariance Estimation}
\label{alg:LRKron}
\begin{algorithmic}[1]
\STATE $\mathbf{S} = \mathbf{\Sigma}_{SCM}$, form $\mathbf{S}(i,j)$, $\overline{\mathbf{S}}(i,j)$.
\STATE Initialize $\mathbf{A}_0$ (or $\mathbf{B}_0$) using \eqref{Eq:Werner}, with $\mathbf{A}_0$ s.t. $\|\mathbf{A}_0\|_F = 1$ (correspondingly $\mathbf{B}_0$).
\WHILE{Objective $\| \mathbf{S}-{ \mathbf{A}_k}\otimes{\mathbf{B}_k}\|_F^2$ not converged}
\STATE $\mathbf{R}_B= \frac{\sum_{i,j}^p a^*_{k,ij}\overline{\mathbf{S}}(i,j)}{\|\mathbf{A}_k\|_F^2}$
\STATE ${\mathbf{B}_{k+1}} = \mathrm{EIG}_{r_b}(\mathbf{R}_B)$
\STATE $\mathbf{R}_A= \frac{\sum_{i,j}^q b^*_{k+1,ij}\mathbf{S}(i,j)}{\|\mathbf{B}_{k+1}\|_F^2}$
\STATE $\mathbf{A}_{k+1} = \mathrm{EIG}_{r_a}(\mathbf{R}_A)$
\ENDWHILE
\RETURN $\hat{\mathbf{A}} = {\mathbf{A}_k},\hat{\mathbf{B}} = {\mathbf{B}_k}$.
\end{algorithmic}
\end{algorithm}



\subsection{Robustness Benefits}
\label{Sec:Robust}
Besides reducing the number of parameters, Kronecker STAP enjoys several other benefits arising from associated properties of the estimation objective \eqref{Eq:SparseOpt}.

The clutter covariance model \eqref{Eq:KronCov} is low rank, motivating the PCA singular value thresholding approach of classical STAP. This approach, however, is problematic in the Kronecker case because of the way low rank Kronecker factors combine. Specifically, the Kronecker product $\mathbf{A}\otimes \mathbf{B}$ has the SVD \cite{loan1992approximation}
\begin{align}
\mathbf{A}\otimes \mathbf{B}=(\mathbf{U}_B\otimes \mathbf{U}_B) (\mathbf{S}_A\otimes \mathbf{S}_B) (\mathbf{U}_A^H\otimes \mathbf{U}_B^H)
\end{align}
where $\mathbf{A}= \mathbf{U}_A \mathbf{S}_A \mathbf{U}_A^H$ and $\mathbf{B} = \mathbf{U}_B \mathbf{S}_B \mathbf{U}_B^H$ are the SVDs of $\mathbf{A}$ and $\mathbf{B}$ respectively. The singular values are $s_A^{(i)} s_B^{(j) }, \: \forall i,j $.
As a result, a simple thresholding of singular values is not equivalent to separate thresholding of the singular values of $\mathbf{A}$ and $\mathbf{B}$ and hence won'€™t necessarily adhere to the space vs. time structure.

For example, suppose that the set of training data is corrupted by inclusion of a sparse set of $w$ moving targets. By the model \eqref{Eq:Moving}, the $i$th moving target gives a return (in the appropriate range bin) of the form
\begin{equation}
\mathbf{z}_i  = \alpha_i \mathbf{a}_i \otimes \mathbf{b}_i,
\end{equation}
where $\mathbf{a}_i,\mathbf{b}_i$ are unit norm vectors. 

This results in a sample data covariance created from a set of observations $\mathbf{n}_m$ with $\mathrm{Cov}[\mathbf{n}_m]= \mathbf{\Sigma}$, corrupted by the addition of a set of $w$ rank one terms
\begin{align}
\label{Eq:corrupt}
\mathbf{S} = \left(\frac{1}{n} \sum_{m=1}^{n} \mathbf{n}_m \mathbf{n}_m^H \right) + \frac{1}{n} \sum_{i = 1}^{w} \mathbf{z}_i \mathbf{z}_i^H .
\end{align}

Let $\tilde{\mathbf{S}} = \frac{1}{n} \sum_{m=1}^{n} \mathbf{n}_m \mathbf{n}_m^H$ and $\tilde{\mathbf{T}} = \frac{1}{n} \sum_{i = 1}^{w} \mathbf{z}_i \mathbf{z}_i^H$. Let ${\lambda}_{S,k}$ be the eigenvalues of $\mathbf{\Sigma}_c$, $\lambda_{S,min} = \min_{k} \lambda_{S,k}$, and let $\lambda_{T,max}$ be the maximum eigenvalue of $\tilde{\mathbf{T}}$. Assume that moving targets are indeed in a subspace orthogonal to the clutter subspace. 
If $\lambda_{T,max} > O(\lambda_{S,min})$, performing rank $r$ PCA on $\mathbf{S}$ 
will result in principal components of the moving target term being included in the ``clutter" covariance estimate. 

If the targets are approximately orthogonal to each other (i.e. not coordinated), then $\lambda_{T,max} = O(\frac{1}{n} |\alpha_i|^2)$. Since the smallest eigenvalue of $\mathbf{\Sigma}_c$ is often small, this is the primary reason that classical LR-STAP is easily corrupted by moving targets in the training data \cite{newstadt2013moving,ginolhac2014exploiting}. 

On the other hand, Kron-STAP is significantly more robust to such corruption. Specifically, consider the \emph{rearranged} corrupted sample covariance:
\begin{equation}
\mathcal{R}(\mathbf{S}) = \frac{1}{n} \sum_{m=1}^w \mathrm{vec}(\mathbf{a}_i\mathbf{a}_i^H) \mathrm{vec}( \mathbf{b}_i \mathbf{b}_i^H)^H + \mathcal{R}(\tilde{\mathbf{S}}).
\end{equation}
This also takes the form of a desired sample covariance plus a set of rank one terms. For simplicity, we ignore the rank constraints in the LR-Kron estimator, in which case we have \eqref{Eq:Werner}
\begin{equation}
\hat{\mathbf{A}}\otimes \hat{\mathbf{B}} = \mathcal{R}^{-1}(\hat{\sigma}_1 \mathbf{u}_1 \mathbf{v}_1^H),
\end{equation}
where $\hat{\sigma}_1 \mathbf{u}_1 \mathbf{v}_1^H$ is the first singular component of $\mathcal{R}(\mathbf{S})$. Let ${\sigma}_1$ be the largest singular value of $\mathcal{R}(\tilde{\mathbf{S}})$. 
The largest singular value $\hat{\sigma}_1$ will correspond to the moving target term only if the largest singular value of $\frac{1}{n} \sum_{m=1}^w \mathrm{vec}(\mathbf{a}_i\mathbf{a}_i^H) \mathrm{vec}( \mathbf{b}_i \mathbf{b}_i^H)^H$ is greater than $O(\sigma_1)$. If the moving targets are uncoordinated, this holds if for some $i$, $\frac{1}{n} |\alpha_i|^2  > O({\sigma}_1)$.
Since $\sigma_1$ models the entire clutter covariance, it is on the order of the total clutter energy, i.e. $\sigma_1^2 = O(\sum_{k=1}^r \lambda_{S,k}^2) \gg \lambda_{S,min}^2$. In this sense Kron-STAP is much more robust to moving targets in training than is LR-STAP. 

\subsection{Kronecker STAP Filters}
\label{Sec:STAP}

Once the low rank Kronecker clutter covariance has been estimated using Algorithm \ref{alg:LRKron}, it remains to identify a filter $\mathbf F$, analogous to \eqref{Eq:RegStap}, that uses the estimated Kronecker covariance model. If we restrict ourselves to subspace projection filters and make the common assumption that the target component in \eqref{Eq:decomp} is orthogonal to the true clutter subspace, then the optimal approach in terms of SINR is to project away the clutter subspace, along with any other subspaces in which targets are not present. If only target orthogonality to the joint spatio-temporal clutter subspace is assumed, then the classical low-rank STAP filter is the projection matrix: 
\begin{align}
\label{Eq:KronUS}
\mathbf{F}_{classical} = \mathbf{I} - \mathbf{U}_A \mathbf{U}_A^H\otimes {\mathbf{U}_B \mathbf{U}_B^H},
\end{align}
where $\mathbf{U}_A, \mathbf{U}_B$ are orthogonal bases for the rank $r_a$ and $r_b$ subspaces of the low rank estimates $\hat{ \mathbf{A}}$ and $\hat {\mathbf{B}}$, respectively, obtained by applying Algorithm \ref{alg:LRKron}. This is the Kronecker product equivalent of the standard STAP projector \eqref{Eq:RegStap}, though it should be noted that \eqref{Eq:KronUS} will require less training data for equivalent performance due to the assumed structure. 

The classical low-rank filter $\mathbf{F} = \mathbf{I} - \mathbf{U}\mathbf{U}^H$ is, as noted in section \ref{SubSec:STAP}, merely an approximation to the SINR optimal filter $\mathbf{F} = \mathbf{\Sigma}^{-1}$. We note, however, that this may not be the only possible approximation. In particular, the inverse of a Kronecker product is the Kronecker product of the inverses, i.e. $\mathbf{A}\otimes\mathbf{B} = \mathbf{A}^{-1} \otimes \mathbf{B}^{-1}$. Hence, we consider using the low rank filter approximation on $\hat{ \mathbf{A}}^{-1}$ and $\hat{\mathbf{B}}^{-1}$ directly. The resulting approximation to $\mathbf{F}_{opt}$ is
\begin{align}
\label{Eq:KronSTAP}
\mathbf{F}_{KSTAP} = (\mathbf{I} - \mathbf{U}_A \mathbf{U}_A^H) \otimes (\mathbf{I} - \mathbf{U}_B \mathbf{U}_B^H) = \mathbf{F}_A \otimes \mathbf{F}_B.
\end{align}
We denote by Kron-STAP the method using LR-Kron to estimate the covariance and \eqref{Eq:KronSTAP} to filter the data. This alternative approximation has significant appeal. Note that it projects away both the spatial and temporal clutter subspaces, instead of only the joint spatio-temporal subspace. This is appealing because by \eqref{Eq:Moving}, no moving target should lie in the same spatial subspace as the clutter, and, as noted in Section \ref{Sec:Model}, if the dimension of the clutter temporal subspace is sufficiently small relative to the dimension $q$ of the entire temporal space, moving targets will have temporal factors ($\mathbf{b}$) whose projection onto the clutter temporal subspace are small. Note that in the event $r_b$ is very close to $q$, either truncating $r_b$ to a smaller value (e.g., determined by cross validation) or setting $\mathbf{U}_B = 0$ is recommended to avoid canceling both clutter and moving targets.

Our clutter model has spatial factor rank $r_a = 1$ \eqref{Eq:KronCov}, implying that the $\mathbf{F}_{KSTAP}$ defined in \eqref{Eq:KronSTAP} projects the array signal $\mathbf x$ onto a $(p-1)(q-r_b)$ dimensional subspace. This is significantly smaller than the $pq-r_b$ dimensional subspace onto which \eqref{Eq:KronUS} and unstructured STAP project the data. As a result, much more of the clutter that ``leaks" outside the primary subspace can be canceled, thus increasing the SINR and allowing lower amplitude moving targets to be detected. 


\subsection{Computational Complexity}
Once the filters are learned, the computational complexity depends on the implementation of the filter and does not depend on the covariance estimation method that determined the filter. 

The computational complexity of learning the LR-STAP filter is dominated by the computation of the clutter subspace, which is $O(p^3q^3)$. Our LR-Kron estimator (Algorithm 1) is iterative, with each iteration having $O(p^2q^2) + O(q^3) + O(q^2p^2) + O(p^3) = O(p^2q^2 + p^3 + q^3)$ computations. If the number of iterations needed is small and $p,q$ are large, there will be significant computational gains over LR-STAP.

\begin{CD}
\subsection{Multipass STAP}
\label{Sec:MultiPass}

In surveillance applications, it is often of interest to determine what, if anything, has changed in a scene between a reference time $t_0$ and a later time $t_1$, e.g. disappearance/appearance of parked vehicles, or the appearance of vehicle footprints \cite{newstadt2013moving,bazi2005unsupervised,bovolo2005detail,ranney2006signal}. When SAR is used for such change detection applications, the radar platform will generally fly past the scene and form a ``reference'' image at time $t_0$, and then at time $t_1 > t_0$ fly a path as close as possible to the original and form a new ``mission'' image. These images are then compared and changes detected. However, moving targets will almost always be detected as changes, along with the changes in the stationary scene background \cite{newstadt2013moving}. When changes of background are of primary interest, moving targets may in fact mask changes in the stationary scene due to displacement and smearing. Hence, it is advantageous to identify moving targets in both scenes prior to or parallel to background change detection. In addition, it may be of interest to detect moving targets in the imagery for their own sake \cite{newstadt2013moving}. We thus exploit the additional scene information arising from having two images to better estimate the clutter subspace, and follow STAP with subsequent noncoherent change detection. 

Our Kronecker STAP based change detection approach concatenates the spatial channels of both registered phase histories ($\mathbf{X}_k$), forming a ``$2p$ channel phase history"
\begin{align}
\mathbf{X} = \left[\begin{array}{c}\mathbf{X}_1\\ \mathbf{X}_2\end{array}\right] \in \mathbb{C}^{2p \times q}.
\end{align}
Since two images are involved with potentially different calibration errors, the clutter subspace is of rank 2. Thus, a rank 2 spatial clutter subspace and a low rank temporal subspace are estimated using LRKron and projected away via the KronSTAP filter. This two pass procedure is easily extended to handle multiple ($>2$) passes of the radar sensor.


\end{CD}

\begin{LongerTheorems}
\section{SINR Analysis}
\label{Sec:Pred}

For a STAP filter matrix $\mathbf{F}$ and steering vector $\mathbf{d}$, the data filter vector is \eqref{Eq:GenFilt} $\mathbf{w} = \mathbf{F}\mathbf{d}$ \cite{ginolhac2014exploiting}. With a target return of the form $\mathbf{x}_{target} = \alpha \mathbf{d}$, the filter output is given by \eqref{Eq:Breakdown}, and the SINR by \eqref{Eq:SINRdef}.



Define $\mathrm{SINR}_{max}$ to be the optimal SINR, achieved at $\mathbf{w}_{opt} = \mathbf{F}_{opt} \mathbf{d}$ \eqref{Eq:opt}.

Suppose that the clutter has covariance of the form \eqref{Eq:KronCov}. 
Assume that the target steering vector $\mathbf{d}$ lies outside both the temporal and spatial clutter subspaces as per above and \cite{ginolhac2014exploiting}. Suppose that LR-STAP is set to use $r$ principal components. Suppose further that Kron STAP uses 1 spatial principal component and $r$ temporal components, so that the total number of principal components of LR-STAP and Kron STAP are equivalent. 

Under these assumptions, if $\sigma$ approaches zero the SINR achieved using LR-STAP, Kron STAP or spatial Kron STAP with infinite training samples achieves \cite{ginolhac2014exploiting} $\mathrm{SINR}_{max}$. 

We analyze the asymptotic convergence rates under the finite sample regime. Define the SINR Loss $\rho$ as the loss of performance when using the estimate $\hat{\mathbf{w}} = \hat{\mathbf{F}}\mathbf{d}$ (corresponding to $\mathrm{SINR}_{out}$) as the filter instead of $\mathbf{w}_{opt}$:
\begin{align}
\rho = \frac{\mathrm{SINR}_{out}}{\mathrm{SINR}_{max}}.
\end{align}

Let $\lambda_i$, $i = 1,\dots,pq$ be the eigenvalues of $\mathbf{\Sigma}_c$. Under the Kronecker model, we have
\begin{equation}
{\lambda}_i = \left\{\begin{array}{ll}s_A^{(1)} s_B^{(i)}, & i = 1,\dots, r_b \\ 0 & i > r_b\end{array}\right.
\end{equation}
since $\mathbf{A}$ only has one nonzero singular value.

\begin{theorem}[LR-STAP SINR \cite{ginolhac2014exploiting}]
\label{Thm:LRSTAP}

For large $n$, the expected SINR Loss of LR-STAP is

\begin{align}
E[\rho] = 1-\frac{1}{n}\sum_{i=1}^r\left(\frac{E[\tau^2]\lambda_i + \sigma^2}{E[\tau^2]\lambda_i}\right)^2,
\end{align}
which in the small $\sigma^2$ regime (typical in SAR \cite{ginolhac2014exploiting}) becomes
\begin{align}
E[\rho] \approx 1-\frac{r}{n}
\end{align}
\end{theorem}
Under the Kronecker model we have
\begin{align}
E[\rho] = 1-\frac{1}{n}\sum_{i=1}^{r}\left(\frac{E[\tau^2]s_B^{(i)} + \frac{\sigma^2}{s_A^{(1)}}}{E[\tau^2]s_B^{(i)}}\right)^2.
\end{align}

We now turn to Kron STAP. Note that the Kron STAP filter can be decomposed into a spatial stage (filtering by $\mathbf{F}_{spatial}$) and a temporal stage (filtering by $\mathbf{F}_{temp}$):
\begin{equation}
\mathbf{F}_{KSTAP} = \mathbf{F}_A\otimes \mathbf{F}_B = \mathbf{F}_{spatial} \mathbf{F}_{temp}
\end{equation}
where $\mathbf{F}_{spatial} = \mathbf{F}_A \otimes \mathbf{I}$ and $\mathbf{F}_{temp} = \mathbf{I} \otimes \mathbf{F}_B$ \eqref{Eq:KronSTAP}.
Under the idealized model in this section, either the spatial or the temporal stage is sufficient to project away the clutter subspace. We assume the naive estimator 
\begin{align}
\label{Eq:Spat}
\hat{\mathbf{A}} = \mathrm{EIG}_1\left(\frac{1}{q}\sum_{i} \mathbf{S}(i,i) \right) = \hat{\psi}\hat{\mathbf{h}}\hat{\mathbf{h}}^H
\end{align}
for the spatial subspace $\mathbf{h}$ ($\|\mathbf{h}\|_2=1$). This is equivalent to approximating the sample spatial covariance as rank 1. The analysis of \cite{ginolhac2014exploiting} thus applies with $r=1$ and $n' = nq$, except some of the samples are correlated. Using the Kronecker structure of the covariance it is trivial to show (for the SIRV distribution) that the worst case occurs when all the clutter temporal correlations are all $\pm1$, in which case $\frac{1}{q} \sum_i \mathbf{S}(i,i)$ reduces to an $n$ iid sample SCM with Gaussian noise variance $\sigma^2/q$ and we can directly obtain the following via Theorem \ref{Thm:LRSTAP}
\begin{theorem}[Kron STAP SINR]
\label{Thm:Sp}
For large $n$ and using the estimator \eqref{Eq:Spat}, the expected SINR Loss of Kron STAP using the estimator \eqref{Eq:Spat} for the spatial subspace satisfies
\begin{align}
E[\rho] \geq 1-\frac{1}{n}\left(\frac{E[\tau^2]\psi + \frac{\sigma^2}{q}}{E[\tau^2]\psi}\right)^2 
\end{align}
where $\psi = s_A^{(1)} \frac{\mathrm{trace}(\mathbf{B})}{q}$.

In the small $\sigma^2$ regime this becomes
\begin{align}
E[\rho] \geq 1-\frac{1}{n}.
\end{align}
\end{theorem}
Since by \eqref{Eq:ClutterCov} $r \leq q$, the gains of using Kron STAP can be quite significant.


Finally we consider the case where errors occurred in estimating the spatial covariance, either due to subspace estimation error or to $\mathbf{A}$ having a rank greater than one, e.g., due to small calibration errors. Specifically, suppose the estimated (rank one) spatial subspace is $\tilde{\mathbf{h}}$, giving a Kron STAP spatial filter $\mathbf{F}_{spatial} = (\mathbf{I} - \tilde{\mathbf{h}}\tilde{\mathbf{h}}^H)\otimes \mathbf{I}$. Suppose further that spatial filtering of the data is followed by the temporal filter $\mathbf{F}_{temp}$ based on the temporal subspace $\mathbf{U}_B$ estimated from the training data. Define the SINR loss $\rho_t|\tilde{\mathbf{h}}$ from using an estimate of $\mathbf{U}_B$ as
\begin{align}
\rho_t|\tilde{\mathbf{h}} = \frac{\mathrm{SINR}_{out}}{\mathrm{SINR}_{max}(\tilde{\mathbf{h}})}
\end{align}
where $\mathrm{SINR}_{max}(\tilde{\mathbf{h}})$ is the maximum achievable SINR given that the spatial filter is fixed at $\mathbf{F}_{spatial}=(\mathbf{I} - \tilde{\mathbf{h}}\tilde{\mathbf{h}}^H)\otimes \mathbf{I}$. Then it is shown in Appendix \ref{App:Pf} that the expected SINR Loss of the temporal Kron STAP stage is given by the following theorem. 

\begin{theorem}[Kron STAP (temporal stage) SINR]
\label{Thm:Temporal}
Suppose that a value for the spatial subspace estimate $\tilde{\mathbf{h}}$ and hence $\mathbf{F}_{spatial}$ is fixed. Then for large $n$ and targets with constant Doppler over the integration interval, the SINR loss from using an estimate of $\mathbf{U}_B$ satisfies
\begin{align}
E[\rho_t | &\tilde{\mathbf{h}}] =\\\nonumber& 1-\frac{\kappa}{n}\sum_{i=1}^{r_b}\left(\frac{(E[\tau^2]s_B^{(i)} + \frac{\sigma^2}{\tilde{\mathbf{h}}^H \mathbf{A}\tilde{\mathbf{h}}})(E[\tau^2]s_B^{(i)} + \frac{\sigma^2}{\kappa\tilde{\mathbf{h}}^H \mathbf{A}\tilde{\mathbf{h}}})}{(E[\tau^2]s_B^{(i)})^2}\right)\\\nonumber
&\quad \quad \quad \qquad \kappa = \frac{\tilde{\mathbf{d}}^H \mathbf{A}\tilde{\mathbf{d}}}{\tilde{\mathbf{h}}^H \mathbf{A}\tilde{\mathbf{h}}}.
\end{align}

In the small $\sigma^2$ regime this becomes
\begin{align}
E[\rho_t | \tilde{\mathbf{h}}] \approx 1-\frac{\kappa r_b}{n}.
\end{align}
\end{theorem}
Note that in the $n \gg p$ regime relevant when $q \gg p$, $\tilde{\mathbf{h}} \approx \mathbf{h}$, where $\mathbf{h}$ is the first singular vector of $\mathbf{A}$. This gives $\tilde{\mathbf{h}}^H \mathbf{A}\tilde{\mathbf{h}} \approx s_A^{(1)}$ and $\kappa \rightarrow 0$ if $\mathbf{A}$ is indeed rank one.  
To avoid cancelation of the moving targets, it is necessary that $r_b \ll q$, and since in the ideal large sample regime all the clutter is removed by the temporal stage, $r_b$ can be smaller than $\mathrm{rank}(\mathbf{B})$. Hence this slower convergence rate on a smaller amount of cancelation than the spatial stage (since $\kappa$ should be small) is still faster than that of LR-STAP in general. 


\end{LongerTheorems}


\begin{ShorterTheorems}

\section{SINR Performance}
\label{Sec:Pred}

For a STAP filter matrix $\mathbf{F}$ and steering vector $\mathbf{d}$, the data filter vector is given by \eqref{Eq:GenFilt}: $\mathbf{w} = \mathbf{F}\mathbf{d}$ \cite{ginolhac2014exploiting}. With a target return of the form $\mathbf{x}_{target} = \alpha \mathbf{d}$, the filter output is given by \eqref{Eq:Breakdown}, and the SINR by \eqref{Eq:SINRdef}.



Define $\mathrm{SINR}_{max}$ to be the optimal SINR, achieved at $\mathbf{w}_{opt} = \mathbf{F}_{opt} \mathbf{d}$ \eqref{Eq:opt}.

Suppose that the clutter has covariance of the form \eqref{Eq:KronCov}. 
Assume that the target steering vector $\mathbf{d}$ lies outside both the temporal and spatial clutter subspaces as justified in \cite{ginolhac2014exploiting}. Suppose that LR-STAP is set to use $r$ principal components. Suppose further that Kron STAP uses 1 spatial principal component and $r$ temporal components, so that the total number of principal components of LR-STAP and Kron STAP are equivalent. Under these assumptions, if the noise variance $\sigma^2$ approaches zero the SINR achieved using LR-STAP, Kron STAP or spatial Kron STAP with infinite training samples achieves $\mathrm{SINR}_{max}$ \cite{ginolhac2014exploiting}. 

We analyze the asymptotic convergence rates under the finite sample regime. Define the SINR Loss $\rho$ as the loss of performance induced by using the estimated STAP filter $\hat{\mathbf{w}} = \hat{\mathbf{F}}\mathbf{d}$ instead of $\mathbf{w}_{opt}$:
\begin{align}
\label{Eq:sinrloss}
\rho = \frac{\mathrm{SINR}_{out}}{\mathrm{SINR}_{max}},
\end{align}
where $\mathrm{SINR}_{out}$ is the output signal to interference ratio when using $\hat{\mathbf{w}}$.



It is shown in \cite{ginolhac2014exploiting} that for large $n$ and small $\sigma$, the expected SINR Loss of LR-STAP under the SIRV model \eqref{Eq:decomp} is
\begin{align}
\label{Eq:lrstp}
E[\rho] = 1-\frac{r}{n}.
\end{align}
This approximation is obtained specializing the result in \cite[Prop. 3.1]{ginolhac2014exploiting} to the case of small $\sigma$. 

We now turn to Kron STAP. Note that the Kron STAP filter can be decomposed into a spatial stage (filtering by $\mathbf{F}_{spatial}$) and a temporal stage (filtering by $\mathbf{F}_{temp}$):
\begin{equation}
\mathbf{F}_{KSTAP} = \mathbf{F}_A\otimes \mathbf{F}_B = \mathbf{F}_{spatial} \mathbf{F}_{temp}
\end{equation}
where $\mathbf{F}_{spatial} = \mathbf{F}_A \otimes \mathbf{I}$ and $\mathbf{F}_{temp} = \mathbf{I} \otimes \mathbf{F}_B$ \eqref{Eq:KronSTAP}.
When the clutter covariance fits our SIRV model, either the spatial or the temporal stage is sufficient to project away the clutter subspace. Assume one adopts the naive estimator 
\begin{align}
\label{Eq:Spat}
\hat{\mathbf{A}} = \mathrm{EIG}_1\left(\frac{1}{q}\sum_{i} \mathbf{S}(i,i) \right) = \hat{\psi}\hat{\mathbf{h}}\hat{\mathbf{h}}^H
\end{align}
for the spatial subspace $\mathbf{h}$ ($\|\mathbf{h}\|_2=1$). 
For large $n$ and small $\sigma$, the expected SINR Loss of Kron STAP using the estimator \eqref{Eq:Spat} for the spatial subspace is given by
%
\begin{align}
\label{Eq:sinrK}
E[\rho] = 1-\frac{1}{n}.
\end{align}
This result is established in \cite[Theorem IV.2]{greenewald2015kronecker} for the SIRV model. The proof is based on applying the LR-STAP result \eqref{Eq:lrstp} to an equivalent rank-one estimator. 
Since by \eqref{Eq:ClutterCov} the full clutter covariance has rank $r \sim q$, the gains of using Kron STAP over LR-STAP (which decays linearly with $r$) can be quite significant.


Next we establish the robustness of the proposed Kron STAP algorithm to estimation errors, for which the SINR loss \eqref{Eq:sinrloss} can only be empirically estimated from training data. Specifically, consider the case where the spatial covariance has estimation errors, either due to subspace estimation error or to $\mathbf{A}$ having a rank greater than one, e.g., due to spatially varying calibration errors. Specifically, suppose the estimated (rank one) spatial subspace is $\tilde{\mathbf{h}}$, giving a Kron STAP spatial filter $\mathbf{F}_{spatial} = (\mathbf{I} - \tilde{\mathbf{h}}\tilde{\mathbf{h}}^H)\otimes \mathbf{I}$. Suppose further that spatial filtering of the data is followed by the temporal filter $\mathbf{F}_{temp}$ based on the temporal subspace $\mathbf{U}_B$ estimated from the training data. Define the resulting SINR loss $\rho_t|\tilde{\mathbf{h}}$ as 
\begin{align}
\rho_t|\tilde{\mathbf{h}} = \frac{\mathrm{SINR}_{out}}{\mathrm{SINR}_{max}(\tilde{\mathbf{h}})}
\end{align}
where $\mathrm{SINR}_{max}(\tilde{\mathbf{h}})$ is the maximum achievable SINR given that the spatial filter is fixed at $\mathbf{F}_{spatial}=(\mathbf{I} - \tilde{\mathbf{h}}\tilde{\mathbf{h}}^H)\otimes \mathbf{I}$. 

We then can obtain the following. 
Suppose that a value for the spatial subspace estimate $\tilde{\mathbf{h}}$ (with $\|\tilde{\mathbf{h}}\|_2=1$) and hence $\mathbf{F}_{spatial}$ is fixed. Let the steering vector for a constant Doppler target be $\mathbf{d} = \mathbf{d}_A \otimes \mathbf{d}_B$ per \eqref{Eq:Moving}, and suppose that $\mathbf{d}_A$ is fixed and $\mathbf{d}_B$ is arbitrary. Then for large $n$ and small $\sigma$, the SINR loss from using an estimate of $\mathbf{U}_B$ follows
\begin{align}
\label{Eq:RhoPart}
E[\rho_t | \tilde{\mathbf{h}}] \approx 1-\frac{\kappa r_b}{n}, \qquad \kappa = \frac{\tilde{\mathbf{d}}^H_A \mathbf{A}\tilde{\mathbf{d}}_A}{\tilde{\mathbf{h}}^H \mathbf{A}\tilde{\mathbf{h}}}.
\end{align}
where $\tilde{\mathbf{d}}_A =\frac{(\mathbf{I} - \tilde{\mathbf{h}} \tilde{\mathbf{h}}^H)\mathbf{d}_A}{\|(\mathbf{I} - \tilde{\mathbf{h}} \tilde{\mathbf{h}}^H)\mathbf{d}_A\|_2}$ and the data follows the SIRV model. The quantity $\kappa$ is the ratio of the clutter energy residual after filtering with $\tilde{\mathbf h}$ normalized by the energy of the clutter canceled by $\tilde{\mathbf h}$, and creates the gap between the SINR at the starting point (only spatial filtering by $\tilde{\mathbf h}$) and the converged result (as $n \rightarrow \infty$, $\rho_t| \tilde{\mathbf{h}} \rightarrow 1$).
A proof sketch of this result is in Appendix \ref{App:Pf}, and more details are given in \cite[Theorem IV.3]{greenewald2015kronecker}.

Note that in the $n \gg p$ regime (relevant when $q \gg p$), $\tilde{\mathbf{h}} \approx \mathbf{h}$, where $\mathbf{h}$ is the first singular vector of $\mathbf{A}$. This gives $\tilde{\mathbf{h}}^H \mathbf{A}\tilde{\mathbf{h}} \approx s_A^{(1)}$ and $\kappa \rightarrow 0$ if $\mathbf{A}$ is indeed rank one. Hence, $\kappa$ can be interpreted as quantifying the adverse effect of mismatch between $\mathbf{A}$ and its estimate. 
From \eqref{Eq:RhoPart} it is seen that cancelation of the moving targets is avoided when $r_b \ll q$. Furthermore, since in the ideal large sample regime all the clutter is removed by the spatial stage, $r_b$ can be smaller than $\mathrm{rank}(\mathbf{B})$, resulting in higher SINR.

In the next section, we provide empirical finite sample validation of these asymptotic results on robustness of the proposed Kron STAP algorithm.

\end{ShorterTheorems}

\section{Numerical Results}
\label{Sec:Results}
\subsection{Dataset}



For evaluation of the proposed Kron STAP methods, we use measured data from the 2006 Gotcha SAR GMTI sensor collection \cite{GotchaData}. This dataset consists of SAR passes through a circular path around a small scene containing various moving and stationary civilian vehicles. The example images shown in the figures are formed using the backprojection algorithm with Blackman-Harris windowing as in \cite{newstadt2013moving}. For our experiments, we use 31 seconds of data, divided into 1 second (2171 pulse) coherent integration intervals.

As there is no ground truth for all targets in the Gotcha imagery, target detection performance cannot be objectively quantified by ROC curves. 
We rely on non ROC measures of performance for the measured data, and use synthetically generated data to show ROC performance gains. In several experiments we do make reference to several higher amplitude example targets in the Gotcha dataset. These were selected by comparing and analyzing the results of the best detection methods available.

\subsection{Simulations}
\begin{figure}[htb]
\centering
\includegraphics[width=2in]{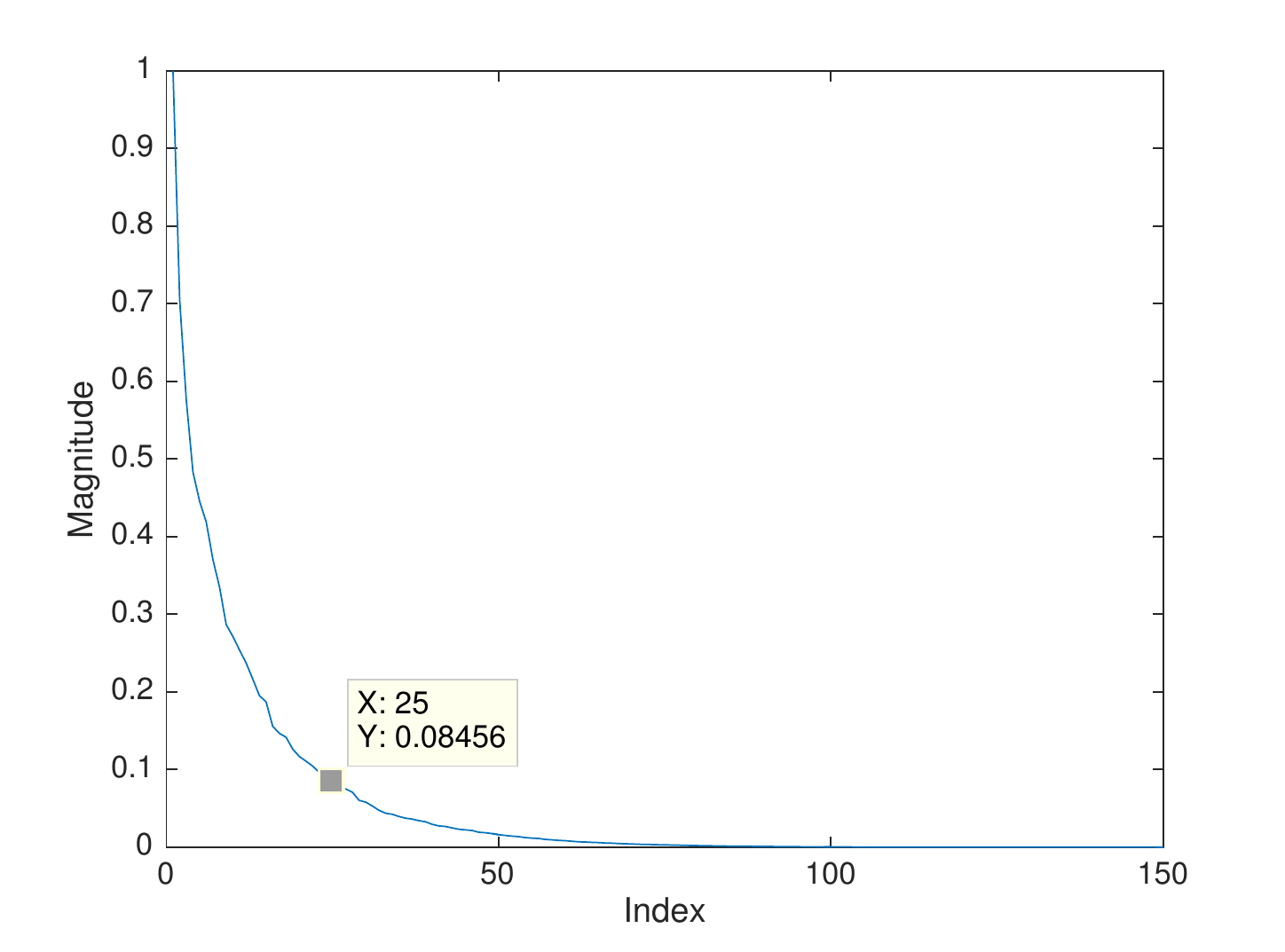}
\caption{Plot of the normalized spectrum of the empirically estimated spatial clutter covariance matrix $\mathbf{B}$, with the subspace dimension $r_b = 25$ chosen as the knee of the curve. We note that our results are not sensitive to small perturbations of $r_b$.}
\label{fig:skree}
\end{figure}

We generated synthetic clutter plus additive noise samples having a low rank Kronecker product covariance. The covariance we use to generate the synthetic clutter via the SIRV model was learned from a set of example range bins extracted from the Gotcha dataset, letting the SIRV scale parameter $\tau^2$ in \eqref{Eq:Cov} follow a chi-square distribution. We use $p=3$, $q=150$, $r_b= 25$, and $r_a = 1$, and generate both $n$ training samples and a set of testing samples. The rank of the left Kronecker factor $\mathbf A$, $r_a$, is 1 as dictated by the spatially invariant antenna calibration assumption and we chose $r_b = 25$ based on a scree plot, i.e., $25$ was the location of  the knee of the spectrum of $\mathbf{B}$ (Figure \ref{fig:skree}). Spatio-temporal Kron-STAP, Spatial-only Kron-STAP, and LR-STAP were then used to learn clutter cancelation filters from the training clutter data. 

\begin{figure}[htb]
\centering
\includegraphics[width=3in]{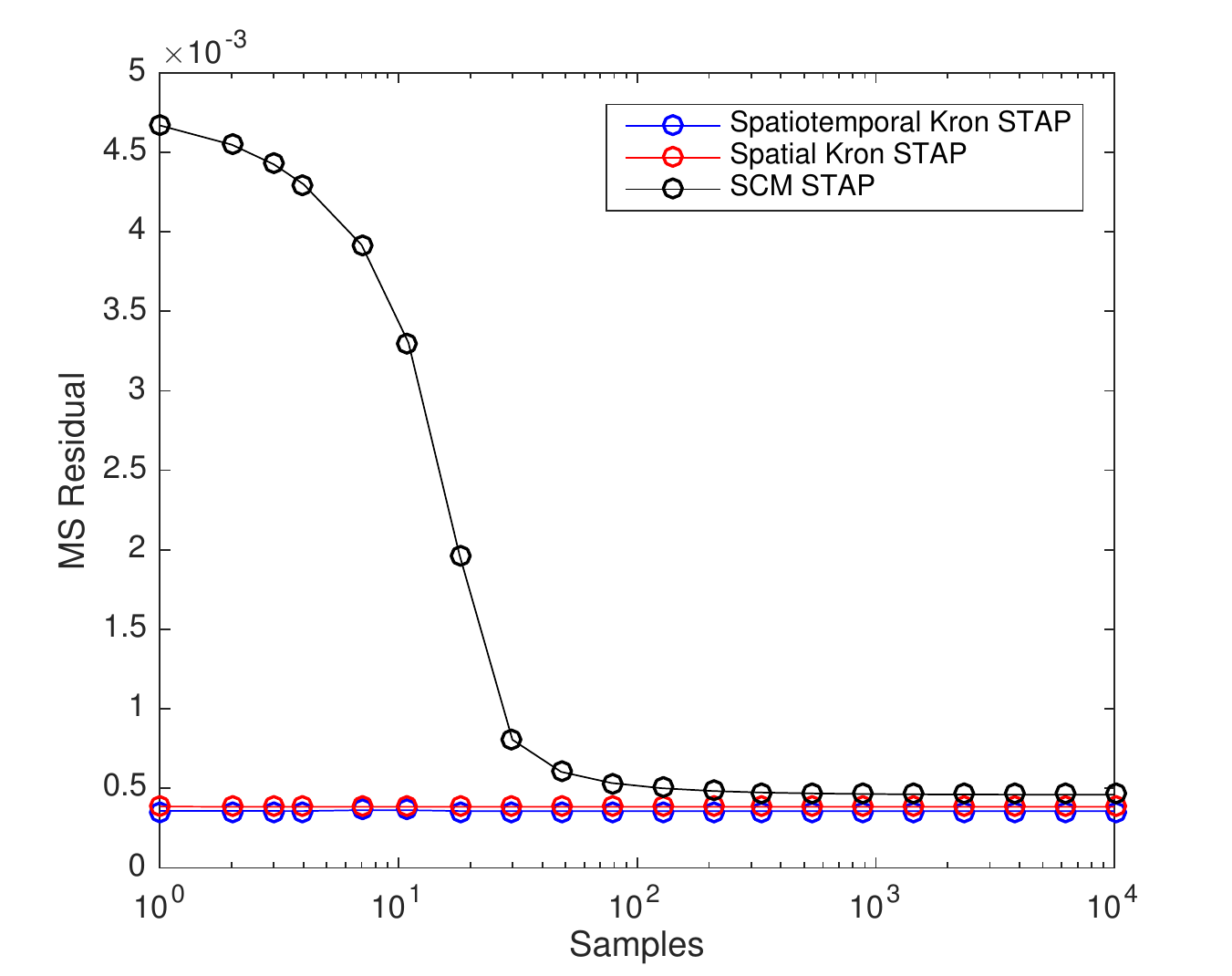}
\includegraphics[width=3in]{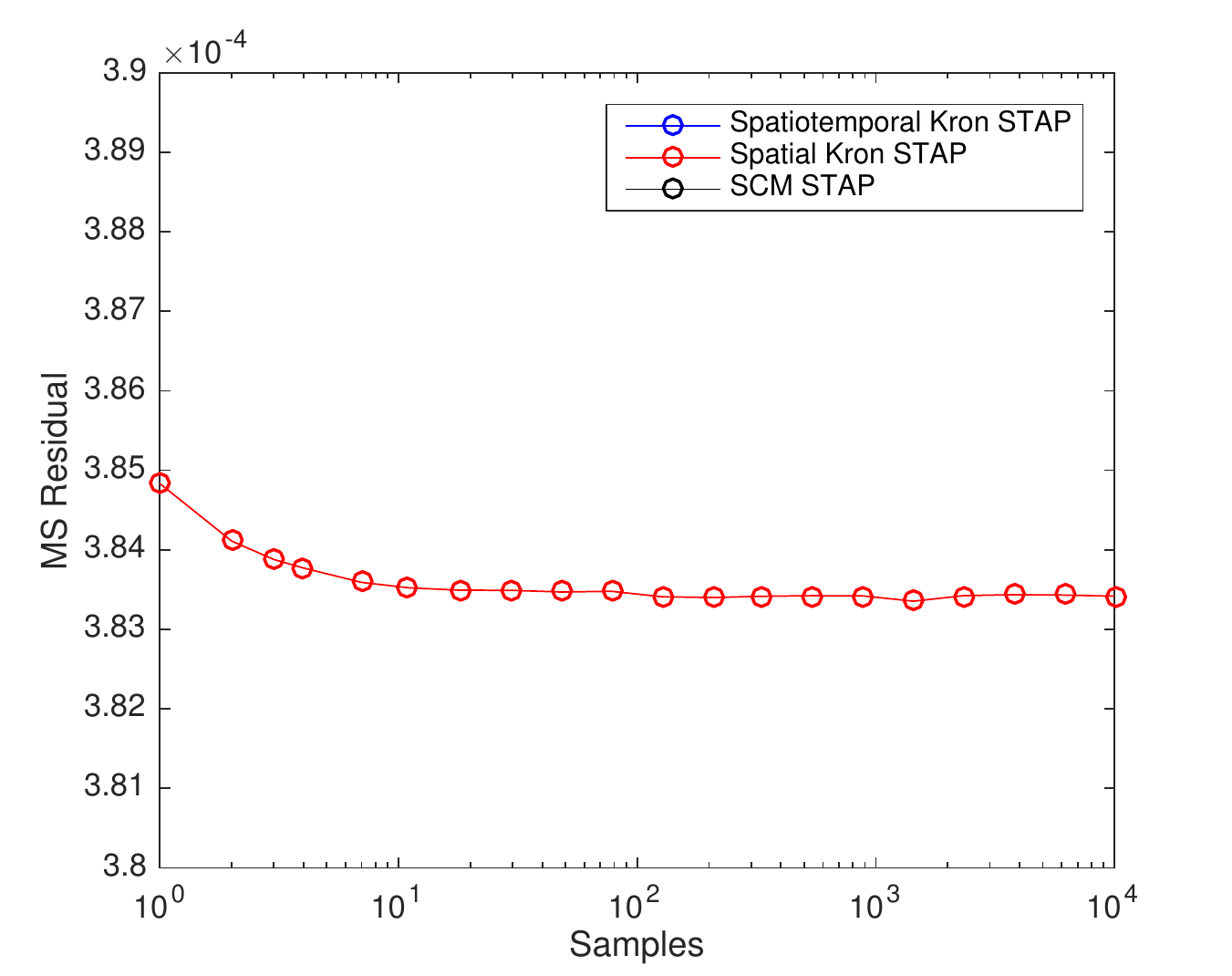}
\caption{Average simulated mean squared residual (MSR), as a function of the number of training samples, of noisy synthetic clutter filtered by spatio-temporal Kron STAP, spatial only Kron STAP, and unstructured LR-STAP (SCM STAP) filters. On the bottom a zoomed in view of a Kron STAP curve is shown. Note the rapid convergence and low MSE of the Kronecker methods.}
\label{Fig:MS}
\end{figure}

The learned filters were then applied to testing clutter data, the mean squared value (MS Residual) of the resulting residual (i.e. $(1/M)\sum_{m=1}^M \|\mathbf{F} \mathbf{x}_m \|_2^2$) was computed, and the result is shown in Figure \ref{Fig:MS} as a function of $n$. Note that the MS Residual corresponds to the average empirical value of the denominator of the SINR \eqref{Eq:SINRdef}, and thus is a target-independent indicator of SINR convergence. The results illustrate the much slower convergence rate of unstructured LR-STAP. as compared to the proposed Kron STAP, which converges after $n=1$ sample. The mean squared residual does not go to zero with increasing training sample size because of the additive noise floor.

\begin{figure}[htb]
\centering
\includegraphics[width=3in]{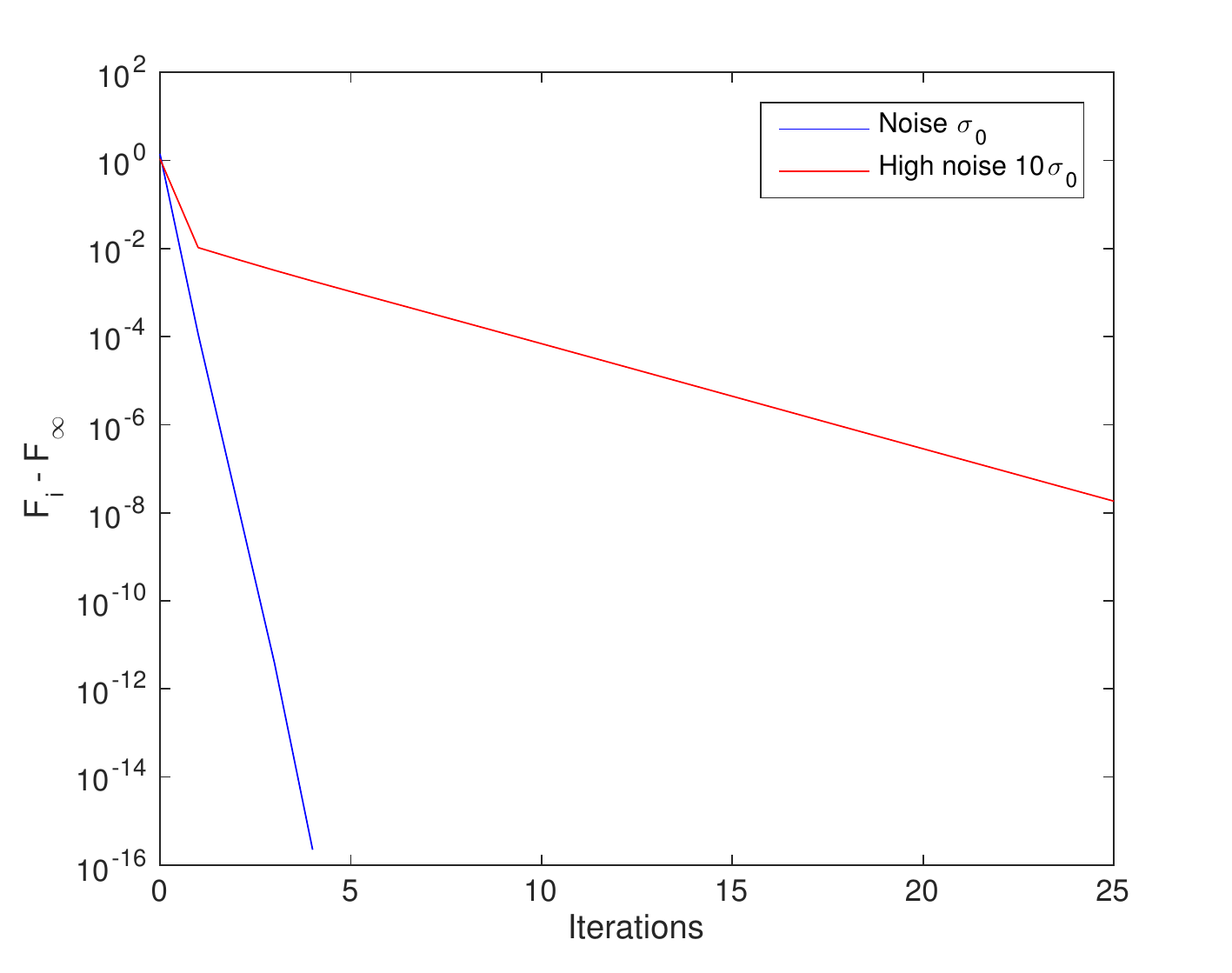}
\caption{Convergence of the LR-Kron algorithm for estimation of the covariance of Figure 1 with $n = 50$. The baseline noise (standard deviation $\sigma_0$) case is shown, along with a high noise example with noise standard deviation $10 \sigma_0$. Shown are logarithmic plots of $F_i - \lim_{i\rightarrow \infty} F_i$ where $F_i = \|\mathbf{S} - \mathbf{A}_i \otimes \mathbf{B}_i\|_F$ as a function of iteration $i$. Note the rapid convergence of the algorithm.}
\label{Fig:Conv}
\end{figure}

As an example of the convergence of Algorithm 1, Figure \ref{Fig:Conv} shows logarithmic plots of $F_i - \lim_{i\rightarrow \infty} F_i$ as a function of iteration $i$, where $F_i = \|\mathbf{S} - \hat{\mathbf{A}}_i \otimes \hat{\mathbf{B}}_i\|_F$. Shown are the results for a sample covariance used in the generation of Figure \ref{Fig:MS} ($n = 50$, noise standard deviation $\sigma_0$), and the results for the case of significantly higher noise (noise standard deviation $10\sigma_0$). The zeroth iteration corresponds to the SVD-based initialization in step 2 of Algorithm 1. In both cases, note the rapid convergence of the algorithm, particularly in the first iteration.

\begin{figure}[htb]
\centering
\includegraphics[width=3in]{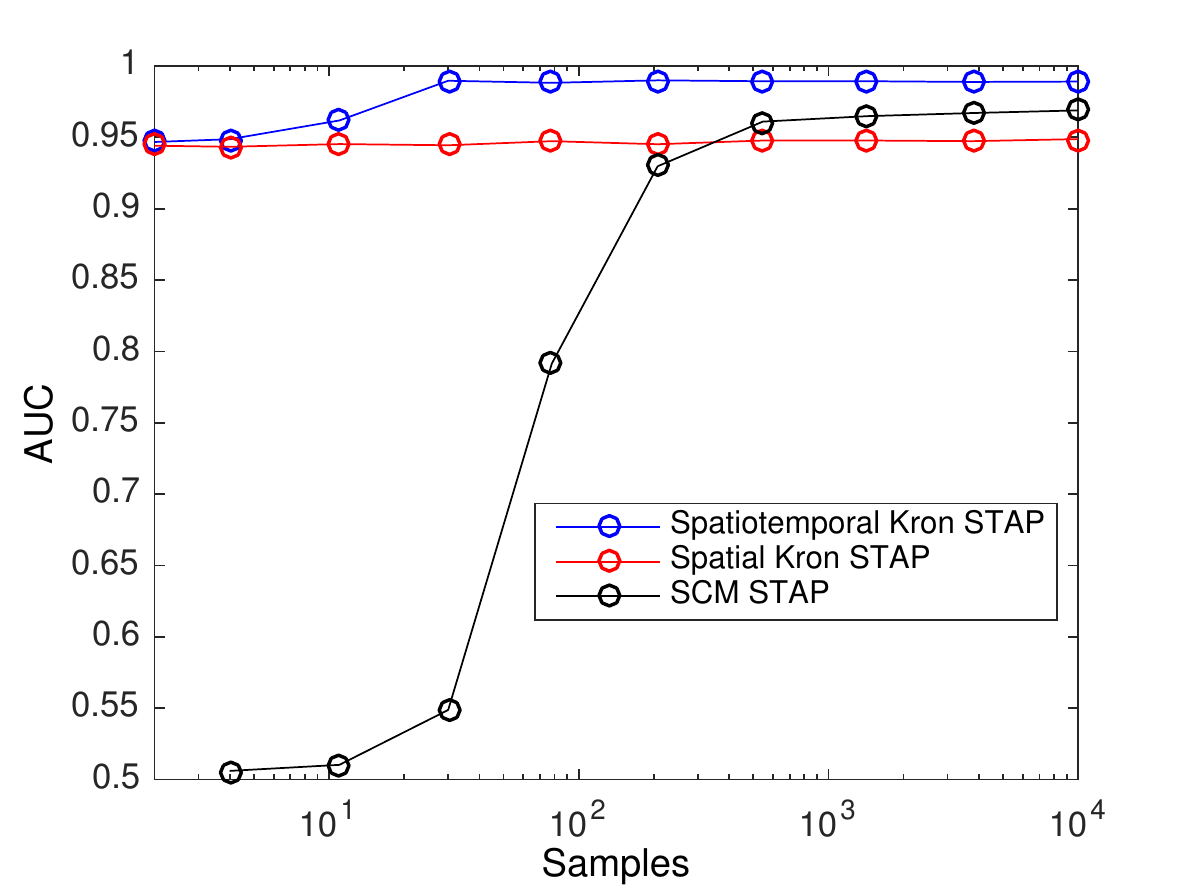}
\caption{Area-under-the-curve (AUC) for the ROC associated with detecting a synthetic target using the steering vector with the largest return, when slight spatial nonidealities exist in the true clutter covariance. 
Note the rapid convergence of the Kronecker methods as a function of the number of training samples, and the superior performance of spatio-temporal Kron STAP to spatial-only Kron STAP when the target's steering vector $\mathbf{d}$ is unknown. }
\label{Fig:ROC}
\end{figure}

To explore the effect of model mismatch due to spatially variant antenna calibration errors ($r_a>1$), we simulated data with a clutter spatial covariance $\mathbf{A}$ having rank 2 with non-zero eigenvalues equal to 1 and $1/30^2$. The STAP algorithms remain the same with $r_a = 1$,  
and synthetic range bins containing both clutter and a moving target are used in testing the effect of this model mismatch on the STAP algorithms. The STAP filter response, maximized over all possible steering vectors, is used as the detection statistic. The AUC of the associated ROC curves is plotted in Figure \ref{Fig:ROC} as a function of the number of training samples. Note again the poor performance and slow convergence of LR-STAP, and that spatio-temporal Kron-STAP converges quickly to the optimal STAP performance, improving on spatial Kron-STAP in a manner consistent with the theoretical result (33). 

\begin{figure}[htb]
\centering
\includegraphics[width=3in]{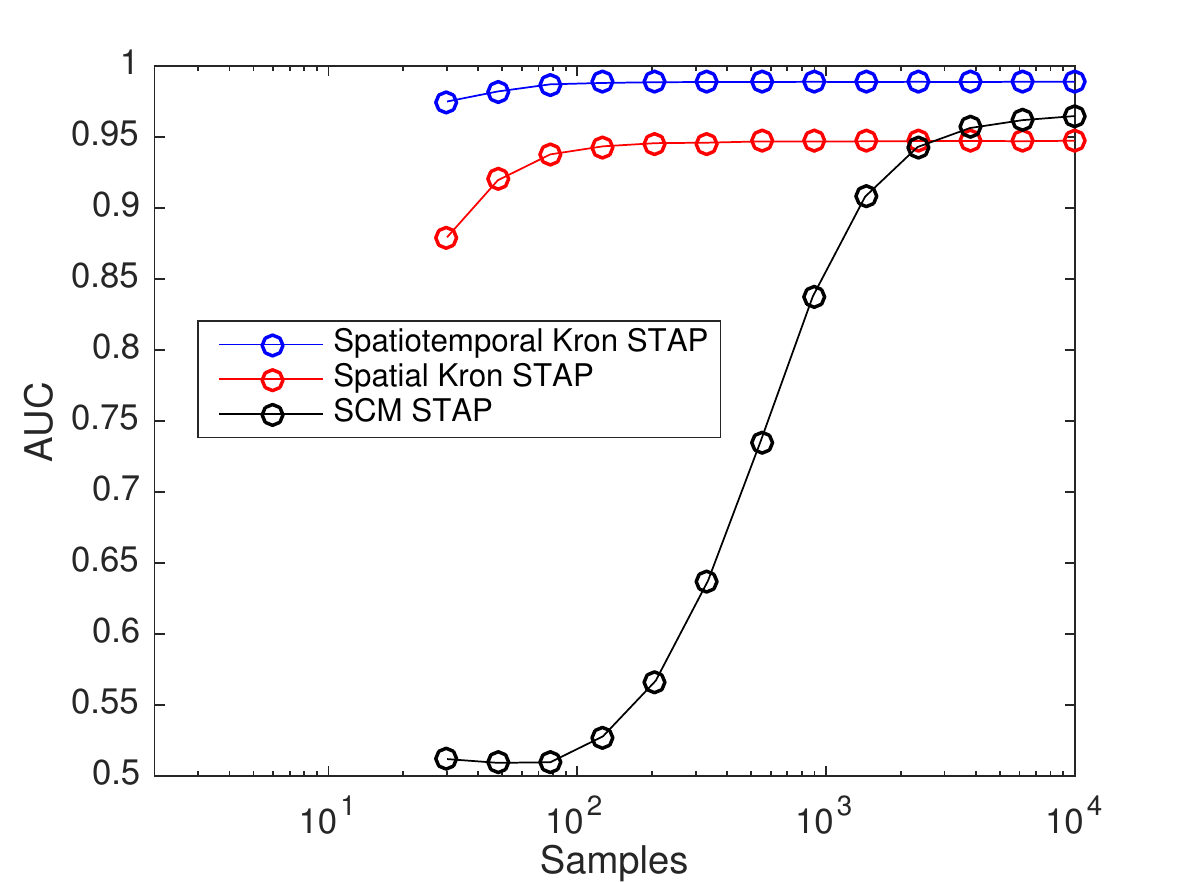}
\caption{Robustness to corrupted training data: AUCs for detecting a synthetic target using the maximum steering vector when (in addition to the spatial nonidealities) 5\% of the training range bins contain targets with random location and velocity in addition to clutter. Note that relative to Figure \ref{Fig:ROC} 
LR-STAP has degraded significantly, whereas the Kronecker methods have not. }
\label{Fig:ROCC}
\end{figure}

Finally, we repeat the AUC vs. sample complexity experiment described in the previous paragraph where 5\% of the training data now have synthetic moving targets with random Doppler shifts. The results are shown in Figure \ref{Fig:ROCC}. 
As predicted by the theory in Subsection \ref{Sec:Robust}, the Kronecker methods remain largely unaffected by the presence of corrupting targets in the training data until the very low sample regime, whereas significant losses are sustained by LR-STAP. This confirms the superior robustness of the proposed Kronecker structured covariance used in our Kron STAP method.


\subsection{Gotcha Experimental Data}

In this subsection, STAP is applied to the Gotcha dataset. For each range bin we construct steering vectors $\mathbf{d}_i$ corresponding to 150 cross range pixels. In single antenna SAR imagery, each cross range pixel is a Doppler frequency bin that corresponds to the cross range location for a stationary target visible at that SAR Doppler frequency, possibly complemented by a moving target that appears in the same bin. Let $\mathbf{D}$ be the matrix of steering vectors for all 150 Doppler (cross range) bins in each range bin. Then the SAR images at each antenna are given by $\tilde{\mathbf{x}} = \mathbf{I}\otimes\mathbf{D}^H\mathbf{x}$ and the STAP output for a spatial steering vector $\mathbf{h}$ and temporal steering $\mathbf{d}_i$ (separable as noted in \eqref{Eq:Moving}) is the scalar
\begin{align}
y_i(\mathbf{h}) = (\mathbf{h}\otimes \mathbf{d}_i)^H \mathbf{F} \mathbf{x}
\end{align}
Due to their high dimensionality, plots for all values of $\mathbf{h}$ and $i$ cannot be shown. Hence, for interpretability we produce images where for each range bin the $i$th pixel is set as $\max_{\mathbf{h}} |y_i(\mathbf{h})|$. More sophisticated detection techniques could invoke priors on $\mathbf{h}$, but we leave this for future work.

\begin{figure*}[]
\begin{centering}
\includegraphics[width=6.6in]{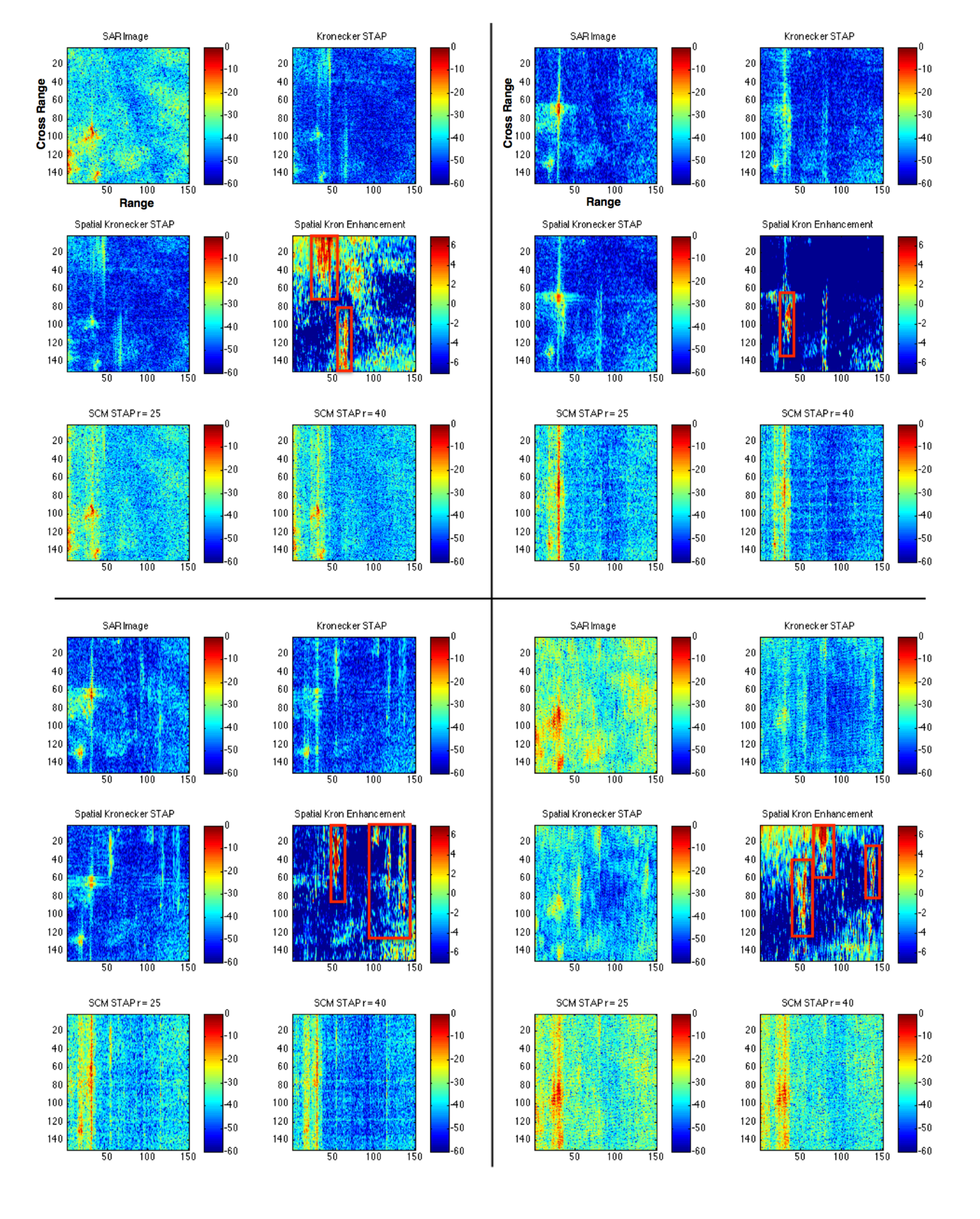}
\end{centering}
\caption{Four example radar images from the Gotcha dataset along with associated STAP results. The lower right example uses 526 pulses, the remaining three use 2171 pulses. Several moving targets are highlighted in red in the spatial Kronecker enhancement plots. Note the superiority of the Kronecker methods. Used Gotcha dataset ``mission" pass, starting times: upper left, 53 sec.; upper right, 69 sec.; lower left, 72 sec.; lower right 57.25 sec.} 
\label{Fig:Examples}
\end{figure*}

Shown in Figure \ref{Fig:Examples} are results for several examplar SAR frames, showing for each example the original SAR (single antenna) image, the results of spatio-temporal Kronecker STAP, the results of Kronecker STAP with spatial filter only, the amount of enhancement (smoothed dB difference between STAP image and original) at each pixel of the spatial only Kronecker STAP, standard unstructured STAP with $r=25$ (similar rank to Kronecker covariance estimate), and standard unstructured STAP with $r = 40$. Note the significantly improved contrast of Kronecker STAP relative to the unstructured methods between moving targets (high amplitude moving targets marked in red in the figure) and the background. Additionally, note that both spatial and temporal filtering achieve significant gains. Due to the lower dimensionality, LR-STAP achieves its best performance for the image with fewer pulses, but still remains inferior to the Kronecker methods. 

\begin{figure}[htb]
\centering
\includegraphics[width=3in]{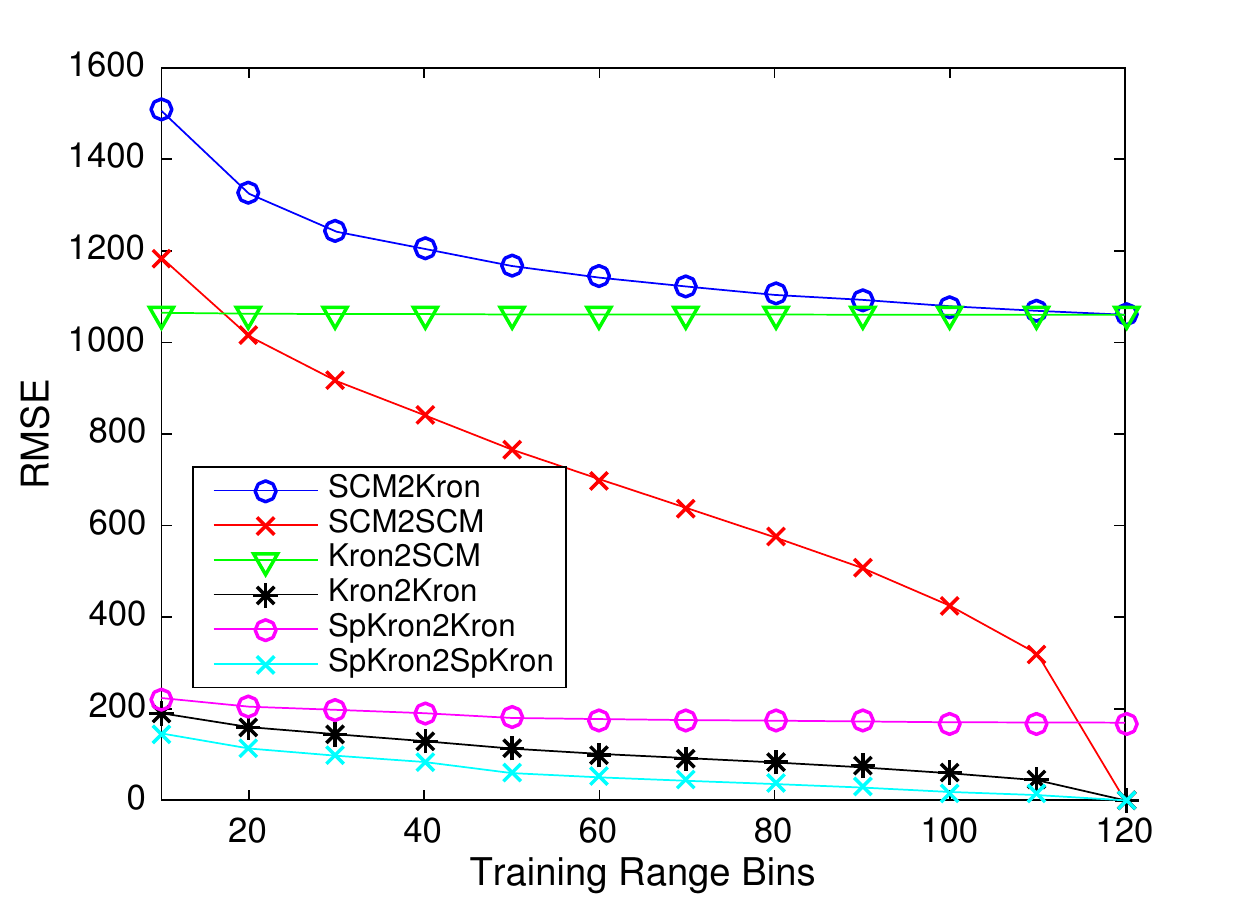}
\includegraphics[width=3in]{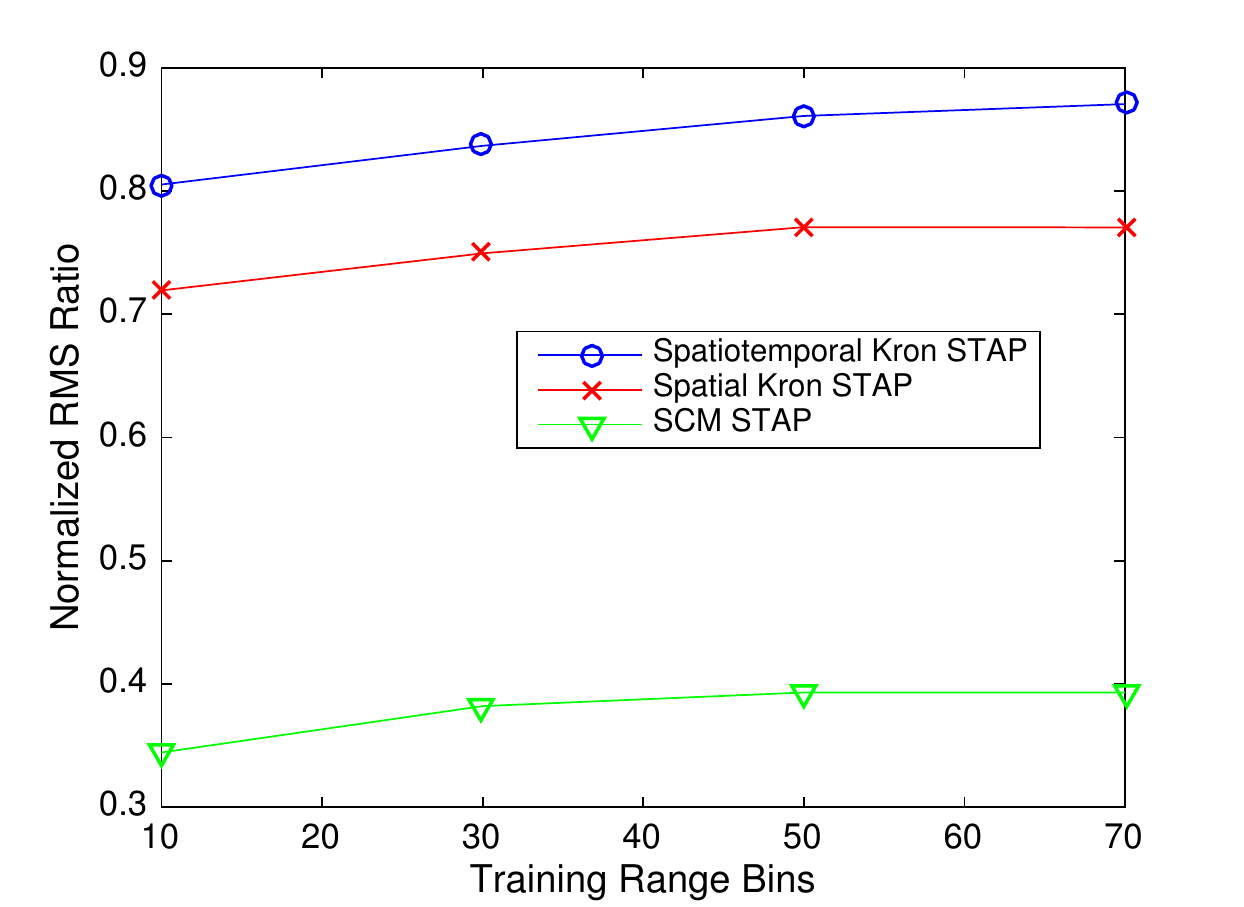}
\caption{Gotcha dataset. Top: Average RMSE of the output of the Kronecker, spatial only Kronecker, and unstructured STAP filters relative to each method's maximum training sample output. Note the rapid convergence and low RMSE of the Kronecker methods. Bottom: Normalized ratio of the RMS magnitude of the brightest pixels in each target relative to the RMS value of the background, for the output of each of Kronecker STAP, spatial Kronecker STAP, and unstructured STAP.}
\label{Fig:RMSE}
\end{figure}

To analyze convergence behavior, a Monte Carlo simulation was conducted where random subsets of the (bright object free) available training set were used to learn the covariance and the corresponding STAP filters. The filters were then used on each of the 31 1-second SAR imaging intervals and the MSE between the results and the STAP results learned using the entire training set were computed (Figure \ref{Fig:RMSE}). 
Note the rapid convergence of the Kronecker methods relative to the SCM based method, as expected. 

Figure \ref{Fig:RMSE} (bottom) shows the normalized ratio of the RMS magnitude of the 10 brightest filter outputs $y_i(\mathbf{h})$ for each ground truthed target to the RMS value of the background, computed for each of the STAP methods as a function of the number of training samples. This measure is large when the contrast of the target to the background is high. The Kronecker methods clearly outperform LR-STAP.

\begin{CD3}
\subsection{Multipass Kron STAP}
Representative two pass Kronecker STAP results are shown in Figure \ref{Fig:CD}, comparing to two pass LR-STAP and to standard (gain calibrated) incoherent change detection. For the STAP methods, noncoherent change detection is performed following filtering by reforming each image (via maximum steering vectors as in the previous section) and subtracting the resulting pixel magnitudes. It can be seen that additional clutter cancelation capabilities can be gained by using Kronecker STAP on multiple passes.

As in the single pass case, Figure \ref{Fig:RMSE_chng} shows relative RMSE convergence results and the normalized RMS ratio between targets and background. Again, Kron STAP outperforms the other methods, and both STAP methods outperform standard incoherent change detection.

\begin{figure}[htb]
\centering
\includegraphics[width=4.6in]{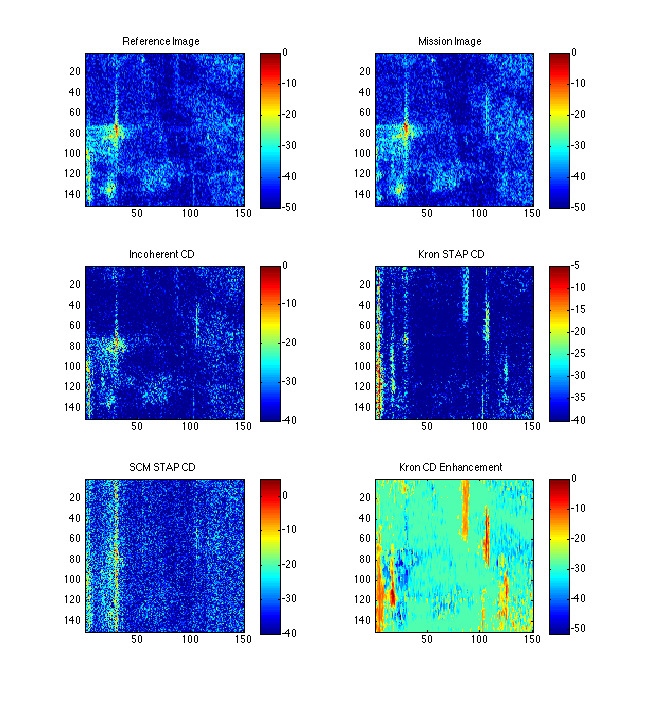}
\caption{Multipass STAP. An example reference and mission image pair are shown, both of which include moving targets. Shown are the results of incoherent change detection, multipass spatio-temporal Kron STAP, multipass LR-STAP, and the multipass spatial Kron STAP enhancement. Note the superior moving target enhancement of the Kronecker methods.}
\label{Fig:CD}
\end{figure}

\begin{figure}[htb]
\centering
\includegraphics[width=2.3in]{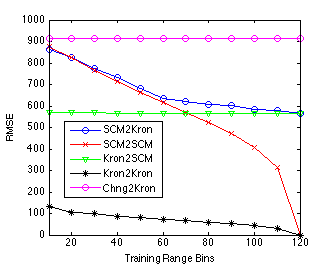}\includegraphics[width=2.3in]{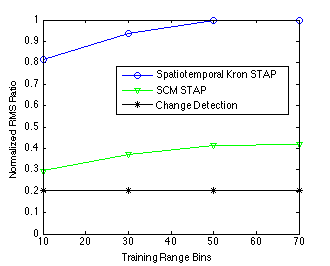}
\caption{Left: Average RMSE of the output of the Kronecker and unstructured STAP filters and incoherent change detection relative to each method's maximum training sample output. Note the rapid convergence and low RMSE of the Kronecker methods. Right: Normalized ratio of the RMS magnitude of the brightest pixels in each target relative to the RMS value of the background, for the output of each of Kronecker STAP, incoherent change detection, and unstructured STAP.}
\label{Fig:RMSE_chng}
\end{figure}
\end{CD3}

\section{Conclusion}
\label{Sec:Conclusion}
In this paper, we proposed a new method for clutter rejection in high resolution multiple antenna synthetic aperture radar systems with the objective of detecting moving targets. Stationary clutter signals in multichannel single-pass radar were shown to have Kronecker product structure where the spatial factor is rank one and the temporal factor is low rank. Exploitation of this structure was achieved using the Low Rank KronPCA covariance estimation algorithm, and a new clutter cancelation filter exploiting the space-time separability of the covariance was proposed. The resulting clutter covariance estimates were applied to STAP clutter cancelation, exhibiting significant detection performance gains relative to existing low rank covariance estimation techniques. As compared to standard unstructured low rank STAP methods, the proposed Kronecker STAP method reduces the number of required training samples and enhances the robustness to corrupted training data. These performance gains were analytically characterized using a SIRV based analysis and experimentally confirmed using simulations and the Gotcha SAR GMTI dataset.

\begin{appendices}
\section{Derivation of Algorithm \ref{alg:LRKron}}
\label{App:LRKron}
We have the following objective function:
\begin{equation}
\label{Eq:SparseOptApp}
\min_{\mathrm{rank}({\mathbf{A}}) = r_a,\mathrm{rank}({\mathbf{B}}) = r_b}\| \mathbf{S}-{ \mathbf{A}}\otimes{\mathbf{B}}\|_F^2.
\end{equation}

To derive the alternating minimization algorithm, fix $\mathbf{B}$ (symmetric) and minimize \eqref{Eq:SparseOptApp} over low rank $\mathbf{A}$:

\begin{align}
\label{Eq:EY}
\arg&\min_{\mathrm{rank}({\mathbf{A}}) = r_a}\| \mathbf{S}-{ \mathbf{A}}\otimes{\mathbf{B}}\|_F^2\nonumber\\\nonumber
=&\arg\min_{\mathrm{rank}({\mathbf{A}}) = r_a}\sum_{i,j}^q \| \mathbf{S}(i,j)-b_{ij}{ \mathbf{A}}\|_F^2\\\nonumber
=&\arg\min_{\mathrm{rank}({\mathbf{A}}) = r_a}\sum_{i,j}^q |b_{ij}|^2\|\mathbf{A}\|_F^2 - 2 \mathrm{Re}[b_{ij} \left\langle \mathbf{A},\mathbf{S}^*(i,j)\right\rangle]\\\nonumber
=&\arg\min_{\mathrm{rank}({\mathbf{A}}) = r_a} \|\mathbf{A}\|_F^2 - 2 \mathrm{Re}\left[\left\langle \mathbf{A},\frac{\sum_{i,j}^q b_{ij}\mathbf{S}^*(i,j)}{\|\mathbf{B}\|_F^2}\right\rangle\right]\\
=&\arg\min_{\mathrm{rank}({\mathbf{A}}) = r_a} \left\|\mathbf{A} - \frac{\sum_{i,j}^q b^*_{ij}\mathbf{S}(i,j)}{\|\mathbf{B}\|_F^2}\right\|_F^2
\end{align}
where $b_{ij}$ is the $i,j$th element of $\hat{\mathbf{B}}$ and $b^*$ denotes the complex conjugate of $b$. This last minimization problem \eqref{Eq:EY} can be solved by the SVD via the Eckart-Young theorem \cite{eckart1936approximation}. First define
\begin{align}
\mathbf{R}_A= \frac{\sum_{i,j}^q b^*_{ij}\mathbf{S}(i,j)}{\|\mathbf{B}\|_F^2},
\end{align}
and let $\mathbf{u}_i^{A},\sigma_i^{A}$ be the eigendecomposition of $\mathbf{R}_A$. The eigenvalues are real and positive because $\mathbf{R}_A$ is positive semidefinite (psd) Hermitian if $\mathbf{B}$ is psd Hermitian \cite{werner2008estimation}. Hence by Eckardt-Young the unique minimizer of the objective \eqref{Eq:EY} is 
\begin{align}
\label{Eq:AofB}
\hat{\mathbf{A}}(\mathbf{B})  =\mathrm{EIG}_{r_a}(\mathbf{R}_A)= \sum_{i = 1}^{r_a} \sigma_i\mathbf{u}_i ^{A} (\mathbf{u}_i^{A})^H.
\end{align}
Note that unless either $\mathbf{S}$ or $\mathbf{B}$ is identically zero, since $\mathbf{B}$ is psd $\mathbf{R}_A$ and hence $\hat{A}(\mathbf{B})$ will be nonzero. 

Similarly, minimizing \eqref{Eq:SparseOptApp} over $\mathbf{B}$ with fixed positive semidefinite Hermitian $\mathbf{A}$ gives the unique minimizer
\begin{align}
\label{Eq:BofA}
\hat{\mathbf{B}}(\mathbf{A})  =\mathrm{EIG}_{r_b}(\mathbf{R}_B) =  \sum_{i = 1}^{r_b} \sigma_i^{B}\mathbf{u}_i^{B} (\mathbf{u}_i^{B})^H,
\end{align}
where now $\mathbf{u}_i^{B}, \sigma_i^{B}$ describes the eigendecomposition of 
\begin{align}
\label{Eq:RB}
\mathbf{R}_B= \frac{\sum_{i,j}^p a^*_{ij}\bar{\mathbf{S}}(i,j)}{\|\mathbf{A}\|_F^2}.
\end{align}
Iterating between computing $\hat{\mathbf{A}}(\mathbf{B})$ and $\hat{\mathbf{B}}(\mathbf{A})$ completes the alternating minimization algorithm. 

By induction, initializing with either a psd Hermitian $\mathbf{A}$ or $\mathbf{B}$ and iterating until convergence will result in an estimate $\hat{\mathbf{A}}\otimes \hat{\mathbf{B}}$ of the covariance that is psd Hermitian since the set of positive semidefinite Hermitian matrices is closed. 

Since for nonzero $\mathbf{S}$ a nonzero $\mathbf{B}_k$ implies a nonzero $\mathbf{A}_{k+1}$ and vice versa, $\mathbf{A}_k$ and $\mathbf{B}_k$ will never go to zero. Hence, the closed-form factorwise minimizers \eqref{Eq:AofB} and \eqref{Eq:BofA} are always uniquely defined, and cannot increase the value of the objective. Thus monotonic convergence of the objective to a value $b$ is ensured \cite{byrne2013alternating}. Since the coordinatewise minimizers are always unique, if \eqref{Eq:AofB} or \eqref{Eq:BofA} result in either $\mathbf{A}_{k+1} \neq \mathbf{A}_k$ or $\mathbf{B}_{k+1} \neq \mathbf{B}_k$ respectively, then the objective function must strictly decrease. Thus, cycles are impossible and $\mathbf{A}_k,\mathbf{B}_k$ must converge to values $\mathbf{A}_*, \mathbf{B}_*$. The value of the objective at that point must be a stationary point by definition, else $\mathbf{A}_*,\mathbf{B}_*$ would not be coordinatewise minima. 


\section{Proof Sketch of \eqref{Eq:RhoPart}}
\label{App:Pf}
This is a proof sketch, the full proof can be found in our technical report \cite[Theorem IV.3]{greenewald2015kronecker}.

After the spatial stage of Kron STAP projects away the estimated spatial subspace $\tilde{\mathbf{h}}$ ( $\|\tilde{\mathbf{h}}\|_2=1$) the remaining clutter has a covariance given by
\begin{align}
((\mathbf{I} - \tilde{\mathbf{h}}\tilde{\mathbf{h}}^H)\mathbf{A}(\mathbf{I} - \tilde{\mathbf{h}}\tilde{\mathbf{h}}^H))\otimes \mathbf{B}.
\end{align}

By \eqref{Eq:Moving}, the steering vector for a (constant Doppler) moving target is of the form $\mathbf{d} = \mathbf{d}_A \otimes \mathbf{d}_B$. Hence, the filtered output is 
\begin{align}
y &= \mathbf{w}^H \mathbf{x} = \mathbf{d}^H\mathbf{F} \mathbf{x}\\\nonumber
& = (\mathbf{d}_A^H \otimes \mathbf{d}_B^H) (\mathbf{F}_A \otimes \mathbf{F}_B)\mathbf{x} \\\nonumber &= ((\mathbf{d}_A^H \mathbf{F}_A)\otimes (\mathbf{d}_B^H \mathbf{F}_B)) \mathbf{x} \\\nonumber &= \mathbf{d}_B^H \mathbf{F}_B \left(\left(\mathbf{d}_A^H \left(\mathbf{I} - \tilde{\mathbf{h}} \tilde{\mathbf{h}}^H\right)\right)\otimes \mathbf{I}\right)\mathbf{x}
\end{align}
Let $\tilde{\mathbf{d}}_A =(\mathbf{I} - \tilde{\mathbf{h}} \tilde{\mathbf{h}}^H)\mathbf{d}_A$  and define $\tilde{\mathbf{c}} = \left(\tilde{\mathbf{d}}_A^H \otimes \mathbf{I}\right)\mathbf{c}$. Then 
\begin{align}
y = \mathbf{d}_B^H \mathbf{F}_B (\tau \tilde{\mathbf{c}} + \tilde{\mathbf{n}}),
\end{align}
where $\tilde{\mathbf{n}} = (\tilde{\mathbf{d}}_A \otimes \mathbf{I})\mathbf{n}$ and
\begin{align}
\mathrm{Cov}[\tilde{\mathbf{c}}] =& (\tilde{\mathbf{d}}_A^H \mathbf{A} \tilde{\mathbf{d}}_A) \mathbf{B} \\\nonumber
\mathrm{Cov}[\tilde{\mathbf{n}}] =& \sigma^2 \mathbf{I},
\end{align}
which are proportional to $\mathbf{B}$ and $\mathbf{I}$ respectively. The scalar $(\tilde{\mathbf{d}}_A^H \mathbf{A} \tilde{\mathbf{d}}_A)$ is small if $\mathbf{A}$ is accurately estimated, hence improving the SINR but not affecting the SINR loss.
Thus, the temporal stage of Kron STAP is equivalent to single channel LR-STAP with clutter covariance $(\tilde{\mathbf{d}}_A^H \mathbf{A} \tilde{\mathbf{d}}_A)\mathbf{B}$ and noise variance $\sigma^2$.






Given a fixed $\hat{\mathbf A} = \tilde{\mathbf{h}}\tilde{\mathbf{h}}^H$, Algorithm \ref{alg:LRKron} dictates \eqref{Eq:RB}, \eqref{Eq:BofA} that 
\begin{align}
\mathbf{R}_B= \sum_{i,j}^p \tilde{h}^*_{i}\tilde{h}^*_{j}\bar{\mathbf{S}}(i,j)\\\nonumber
\hat{\mathbf{B}} = \mathrm{EIG}_{r_b}(\mathbf{R}_B),
\end{align}
which is thus the low rank approximation of the sample covariance of 
\begin{align}
\mathbf{x}_h = \mathbf{x}_{c,h} + \mathbf{n}_h =  (\tilde{\mathbf{h}}\otimes \mathbf I)^H (\mathbf{x}_c + \mathbf{n}).
\end{align}
Since $\mathbf{x}_c = \tau \mathbf{c}$, $\mathbf{x}_{c,h} = \tau (\tilde{\mathbf{h}}\otimes \mathbf I)^H \mathbf{c}$ is an SIRV (Gaussian random vector $(\tilde{\mathbf{h}}\otimes \mathbf I)^H \mathbf{c}$ scaled by $\tau$) with 
\begin{equation}
\mathrm{Cov}[\mathbf{x}_{c,h}] = \tau^2 (\tilde{\mathbf{h}}^H \mathbf{A} \tilde{\mathbf{h}})\mathbf{B}
\end{equation}
Furthermore, $\mathbf{n}_{h} =(\tilde{\mathbf{h}}\otimes \mathbf I)^H \mathbf{n}$ which is Gaussian with covariance $\sigma^2 \mathbf{I}$. Thus, in both training and filtering the temporal stage of Kron STAP is exactly equivalent to single channel LR STAP. Hence we can directly apply the methods used to prove the bound for LR STAP, which after some work results in \eqref{Eq:RhoPart} as desired. 


\end{appendices}
\bibliographystyle{IEEETranS}
\bibliography{CAMSAP_bib}

\begin{IEEEbiography}[{\includegraphics[width=1in,height=1.25in,clip,keepaspectratio]{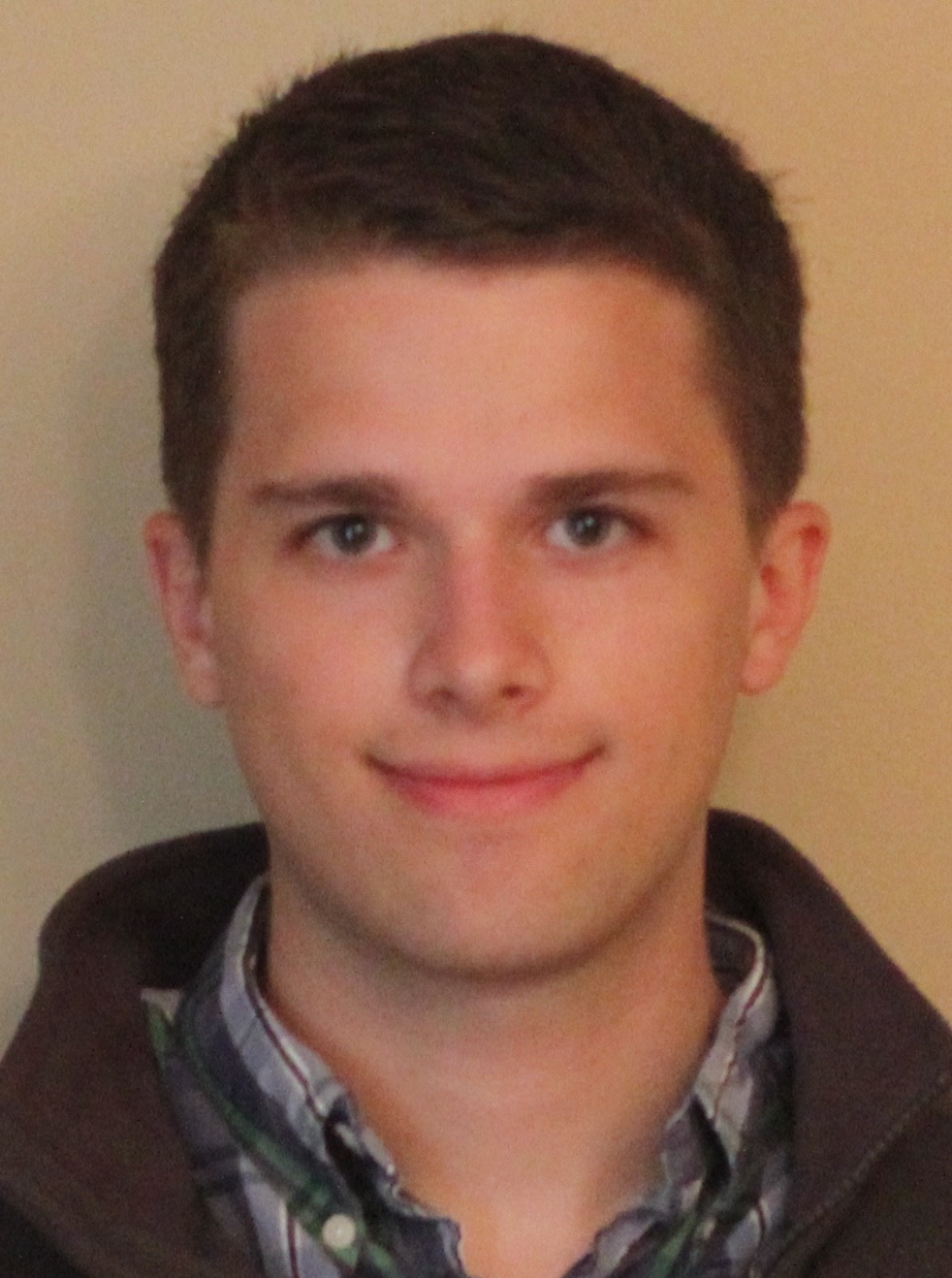}}]{Kristjan Greenewald}
(S'11) Kristjan Greenewald received his B.S. (magna cum laude) and M.S. degrees in Electrical Engineering from Wright State University, in 2011 and 2012 respectively. He is currently pursuing the Ph.D. degree in Electrical Engineering and Computer Science at the University of Michigan, Ann Arbor. He has had research internships at the US Air Force Research Laboratory (2012, 2013, 2014) and at M.I.T. Lincoln Laboratory (2015). His research interests include statistical signal processing, machine learning, and computer vision. 
\end{IEEEbiography}

\begin{IEEEbiography}[{\includegraphics[width=1in,height=1.25in,clip,keepaspectratio]{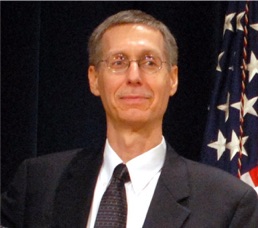}}]{Edmund Zelnio}
Edmund G. Zelnio graduated
from Bradley University, Peoria, Illinois, in 1975 and has pursued
doctoral studies at The Ohio State University in electromagnetics
and at Wright State University in signal processing.
During his 40-year career with the Air Force Research Laboratory
(AFRL), Wright Patterson Air Force Base, Ohio, he has
spent 38 years working in the area of automated exploitation of
imaging sensors primarily addressing synthetic aperture radar.
He is a former division chief and technical advisor of the Automatic
Target Recognition Division of the Sensors Directorate in
AFRL and serves in an advisory capacity to the Department of
Defense and the intelligence community. He is currently the
director of the Automatic Target Recognition Center in AFRL.
He is the recipient of the 53rd Department of Defense Distinguished
Civilian Service Award and is a fellow of the AFRL.
\end{IEEEbiography}

\begin{IEEEbiography}[{\includegraphics[width=1in,height=1.25in]{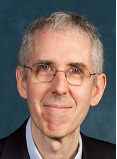}}]{Alfred O. Hero III}

Alfred O. Hero III received the B.S. (summa cum laude) from Boston University (1980) and the Ph.D from Princeton University (1984), both in Electrical Engineering. Since 1984 he has been with the University of Michigan, Ann Arbor, where he is the R. Jamison and Betty Williams Professor of Engineering and co-director of the Michigan Institute for Data Science (MIDAS). His primary appointment is in the Department of Electrical Engineering and Computer Science and he also has appointments, by courtesy, in the Department of Biomedical Engineering and the Department of Statistics. From 2008 to 2013 he held the Digiteo Chaire d'Excellence, sponsored by Digiteo Research Park in Paris, located at the Ecole Superieure d'Electricite, Gif-sur-Yvette, France. He has held other visiting positions at LIDS Massachusetts Institute of Technology (2006), Boston University (2006), I3S University of Nice, Sophia-Antipolis, France (2001), Ecole Normale Sup\'erieure de Lyon (1999), Ecole Nationale Sup\'erieure des T\'el\'ecommunications, Paris (1999), Lucent Bell Laboratories (1999), Scientific Research Labs of the Ford Motor Company, Dearborn, Michigan (1993), Ecole Nationale Superieure des Techniques Avancees (ENSTA), Ecole Superieure d'Electricite, Paris (1990), and M.I.T. Lincoln Laboratory (1987 - 1989).

Alfred Hero is a Fellow of the Institute of Electrical and Electronics Engineers (IEEE). He received the University of Michigan Distinguished Faculty Achievement Award (2011). He has been plenary and keynote speaker at several workshops and conferences. He has received several best paper awards including: an IEEE Signal Processing Society Best Paper Award (1998), a Best Original Paper Award from the Journal of Flow Cytometry (2008), a Best Magazine Paper Award from the IEEE Signal Processing Society (2010), a SPIE Best Student Paper Award (2011), an IEEE ICASSP Best Student Paper Award (2011), an AISTATS Notable Paper Award (2013), and an IEEE ICIP Best Paper Award (2013). He received an IEEE Signal Processing Society Meritorious Service Award (1998), an IEEE Third Millenium Medal (2000), an IEEE Signal Processing Society Distinguished Lecturership (2002), and an IEEE Signal Processing Society Technical Achievement Award (2014). He was President of the IEEE Signal Processing Society (2006-2007). He was a member of the IEEE TAB Society Review Committee (2009), the IEEE Awards Committee (2010-2011), and served on the Board of Directors of the IEEE (2009-2011) as Director of Division IX (Signals and Applications). He served on the IEEE TAB Nominations and Appointments Committee (2012-2014). Alfred Hero is currently a member of the Big Data Special Interest Group (SIG) of the IEEE Signal Processing Society. Since 2011 he has been a member of the Committee on Applied and Theoretical Statistics (CATS) of the US National Academies of Science.

Alfred Hero's recent research interests are in the data science of high dimensional spatio-temporal data, statistical signal processing, and machine learning. Of particular interest are applications to networks, including social networks, multi-modal sensing and tracking, database indexing and retrieval, imaging, biomedical signal processing, and biomolecular signal processing.

\end{IEEEbiography}

\end{document}